\documentclass[lettersize,journal]{IEEEtran}
\usepackage{amsmath,amsfonts,bm}
\usepackage{amssymb}
\usepackage{algorithmic}
\usepackage{algorithm}
\usepackage{amsthm}
\usepackage{array}
\usepackage[caption=false,font=normalsize,labelfont=sf,textfont=sf]{subfig}
\usepackage{array,multirow,graphicx}
\usepackage{color,soul}
\usepackage{multirow}
\definecolor{mygray}{gray}{.9}
\usepackage{textcomp}
\usepackage{stfloats}
\usepackage{url}
\usepackage{verbatim}
\usepackage{graphicx}
\usepackage{cite}
\usepackage{graphicx}
\usepackage{amsmath}
\usepackage{amssymb}
\usepackage{booktabs}
\usepackage{mathtools}
\usepackage{bm}
\showoutput
\usepackage[table]{xcolor}

\newtheorem*{remark}{Remark}

\newtheorem{lemma}{Lemma}
\newtheorem{proposition}{Proposition}

\newcommand{\defeq}{\mbox {$  \ \stackrel{\Delta}{=} $}}

\usepackage{algorithmic}        
\usepackage{algorithm}  
\usepackage{stfloats}

\begin{document}
\title{Coded Deep Learning: Framework and Algorithm}

\author{En-hui~Yang,~\IEEEmembership{Fellow,~IEEE,} Shayan~Mohajer~Hamidi,~\IEEEmembership{Member,~IEEE.}

\IEEEcompsocitemizethanks{\IEEEcompsocthanksitem E.-H. Yang and S. M. Hamidi are with the Department of Electrical and Computer
Engineering, University of Waterloo, Waterloo, ON N2L 3G1, Canada (e-mail:
smohajer@uwaterloo.ca; ehyang@uwaterloo.ca).
}
}


 
\maketitle
% \IEEEtitleabstractindextext{%
\begin{abstract}
The success of deep learning (DL) is often achieved at the expense of large model sizes and high computational complexity during both training and post-training inferences, making it difficult to train and run large models in a resource-limited environment. To alleviate these issues, this paper introduces a new framework dubbed ``coded deep learning'' (CDL), which integrates information-theoretic coding concepts into the inner workings of DL,  aiming to substantially compress model weights and activations, reduce computational complexity at both training and post-training inference stages, and enable efficient model/data parallelism. Specifically, within CDL, (i) we first propose a novel probabilistic method for quantizing both model weights and activations, and its soft differentiable variant which offers an analytic formula for gradient calculation during training; (ii) both the forward and backward passes during training are executed over quantized weights and activations, which eliminates a majority of floating-point operations and reduces the training computation complexity; (iii) during training, both weights and activations are entropy constrained so that they are compressible in an information-theoretic sense at any stage of training, which in turn reduces communication costs in cases where model/data parallelism is adopted; and (iv) the trained model in CDL is by default in a quantized format with compressible quantized weights,  reducing post-training inference complexity and model storage complexity. Additionally, a variant of CDL, namely relaxed CDL (R-CDL), is presented to further improve the trade-off between validation accuracy and compression at the disadvantage of full precision operation involved in forward and backward passes during training with other advantageous features of CDL intact. Extensive empirical results show that CDL and R-CDL outperform the state-of-the-art algorithms in DNN compression in the literature. 
\end{abstract}


\section{Introduction} \label{sec:intro}
\IEEEPARstart{D}{eep} neural networks (DNNs) have demonstrated remarkable performance across diverse applications, from computer vision \cite{krizhevsky2012imagenet, he2016deep} to natural language processing \cite{kenton2019bert, conneau2019cross}. The exceptional performance of these DNNs is largely attributed to their large model sizes, with high-performance models often containing hundreds of millions or even billions of parameters. However, this poses challenges for deployment in resource-limited environments due to (i) significant storage overhead, and (ii) intensive computational and memory requirements during both training and post-training inferences.

To partially address the above challenges and facilitate the efficient deployment of large DNNs on resource-limited devices, a multitude of model compression techniques have been proposed in the literature. These methods include quantization \cite{lsq,lsq+,whitepaper,9404314}, pruning \cite{liu2017learning,liu2018rethinking,9381660,9758156}, and knowledge distillation \cite{hinton2015distilling,9830618,9839519,ye2024bayes}. Among these techniques, quantization methods have attracted significant attention due to their promising hardware compatibility across various architectures \cite{jouppi2017datacenter,sharma2018bit}.

Quantization can be conducted either during training or after training, with the former called quantization-aware training (QAT) \cite{pact2018, lsq,lsq+}, and the latter called post-training quantization (PTQ) \cite{choukroun2019low}. Generally, PTQ methods often experience significant performance degradation when deployed for low-bit quantization. On the other hand, QAT methods proposed in the literature also face several drawbacks: (i) the non-differentiability of quantization functions used in QAT methods necessitates the use of gradient approximation techniques during training \cite{bengio2013estimating, hinton2012ste}, leading to inferior results;
(ii) since they are often applied to pre-trained full-precision (FP) models\footnote{Although some QAT methods train models from scratch \cite{pact2018,lq-net}, their performance in achieving low-bit quantization is typically inferior compared to methods that start from a pre-trained model (see Section \ref{sec:exp}).} to control performance degradation \cite{nagel2022overcoming,tang2022mixed,lsq,lsq+,apot}, they depend on the availability of pre-trained FP models, which may not always be available in the first place, and also introduce more training computation complexity even if pre-trained FP models are available;  and (iii)  more importantly, they often neglect the cost of weight and activation communication required during training in model and data parallelism that is crucial for handling massive DNNs \cite{narayanan2021efficient}. 

To tackle the issues mentioned above, in this paper we shall take a different approach to DNN compression. Specifically, by introducing information-theoretic coding concepts into the core principles of deep learning (DL), we aim to establish a new training framework, dubbed ``coded deep learning'' (CDL), to meet the following four \textit{criteria} simultaneously: 

\noindent $\bullet$ \textbf{\textit{C1} [High validation accuracy]:} The new framework should 
 preserve or even  improve the classification accuracy achieved by models trained using conventional FP methods. Additionally, it should offer better accuracy-compression trade-offs compared to existing quantization methods in the literature. 

\noindent $\bullet$ \textbf{\textit{C2} [Reduced training and inference complexity]:} In the new framework,  forward and backward passes during training should be executed on quantized weights and activations so that  a majority of floating-point operations (FLOPs) can be replaced with low-precision computations, resulting in, possibly, accelerated training times,  reduced memory usage, lower power consumption, and enhanced area efficiency \cite{zhou2016incremental, gholami2022survey}. In addition, the trained model via the new framework should be, by default, in a quantized format so that the same benefits in terms of memory, power, area efficiency,etc. can be carried over to post-training inferences.

\noindent $\bullet$ \textbf{\textit{C3} [Improved model and data parallelism]:} To train a massive model such as  GPT-3 with 175 billion parameters \cite{narayanan2021efficient,brown2020language}, two different modes of parallelism are often employed: (i) pipeline model parallelism \cite{huang2019gpipe,kosson2021pipelined,yang2021pipemare}, where model layers are distributed across multiple devices, necessitating activation communication between devices; and (ii) data parallelism \cite{xing2015petuum}, where the input dataset is partitioned and distributed among different devices, requiring communication of model weights. To have improved model/data parallelism, in the new framework, weights and activations should be compressible in an information-theoretic sense at any stage of training so that the cost of weight and activation communication in model/data parallelism can be reduced.

\noindent $\bullet$ \textbf{\textit{C4} [Reduced storage complexity]:} In the new framework, the weights of the trained model should be 
 compressible in an information-theoretic sense so that after encoding, they require less storage complexity. 
 
How does CDL meet the aforementioned criteria simultaneously?  The first key step is to introduce novel probabilistic quantizers each using a trainable conditional probability mass function (CPMF), in a way similar to rate distortion theory \cite{berger1971rate, yang1997fixed}. These probabilistic quantizers are then used to quantize both model weights and activations during training, and model activations during post-training inferences to satisfy \textit{C2}. Based on its distortion analysis, each of these probabilistic quantizer has a neat \textit{soft deterministic} differentiable approximation. The latter in turn has a nice analytic formula for computing its partial derivatives with respect to (w.r.t.) its input $\theta$ and the quantization step size $q$. Since each of the probabilistic quantizers and its \textit{soft deterministic} approximation are very close to each other, the analytic formula 
can be utilized  as a proxy for gradient calculation during the training stage. Due to this analytic computation, models trained using CDL achieve validation accuracy surpassing that of QAT methods that primarily rely on coarse approximations of the gradients, thereby satisfying \textit{C1} effectively. %It is worth to mention that the idea behind the probabilistic quantizer and its \textit{soft deterministic} approximation is applicable to non-uniform quantization as well, although our focus in this paper is on uniform quantization. 

The second key step is to incorporate information-theoretic coding ideas into the inner workings of DL to make model weights and activations compressible at any stage of training. To this end, given that in CDL, weights and activations are quantized using the proposed probabilistic quantizers, the marginal probability mass functions (MPMFs) of randomly quantized weights and activations for each layer are calculated, and the entropies of these MPMFs are then constrained during training. In other words, in CDL, both the cross-entropy (CE) loss used in DL and entropies of these MPMFs are minimized jointly. Since the compression rates of quantized weights and activations are roughly equal to the entropies of these MPMFs if entropy coding (such as Huffman coding) is used, by minimizing these entropies during training, we make sure that both weights and activations are compressible in an information-theoretic sense at any stage of training, thereby satisfying both \textit{C3} and  \textit{C4}. 

To further improve the trade-off between validation accuracy and compression, CDL is modified by replacing each of the probabilistic quantizers with its \textit{soft deterministic} approximation, yielding a variant referred to as relaxed-CDL (R-CDL). Except that forward and backward passes are now operated with full precision, all other features of CDL remain there in R-CDL. In particular, weights and activations are still compressible at any stage of training in R-CDL. They are quantized using the newly obtained CPMFs whenever the inter-device weight and activation communication need arises in model/data parallelism; weights are still quantized at the end of training. In comparison with CDL, R-CDL has the advantage that all gradients can now be computed analytically without approximation during training at the expense of FP forward and backward passes. This advantage helps R-CDL deliver better accuracy-compression performance than CDL. 

We have conducted a comprehensive set of experiments on CIFAR-100 \cite{krizhevsky2009learning} and ImageNet datasets \cite{krizhevsky2012imagenet}. It is shown that CDL and R-CDL offer superior performance compared to state-of-the-art (SOTA) QAT methods in the literature. The remainder of this paper is organized as follows. Section \ref{Sec:related} briefly reviews related works. Section \ref{sec:prelem} first introduces the notation to be used throughout the paper, and then discusses the preliminary knowledge based on which CDL is established. In Section \ref{sec:quant}, we present the novel probabilistic quantization method along with its differentiable \textit{soft deterministic} variant. Section \ref{sec:CDL} describes  CDL and R-CDL in details. In addition, extensive experimental results on some datasets along with comparison with SOTA methods are reported in Section \ref{sec:exp}, and finally Section \ref{Sec:conclusion} concludes the paper.

\section{Related Works} \label{Sec:related}
\subsection{Quantization-Aware Training (QAT)}
% Specifically, consider $\theta$ as a parameter, which could be either a weight or activation parameter, that needs to be quantized. Assuming that $q$ is the step-size, the quantized version of $\theta$, denoted by $\hat{\theta}$ is calculated as follows
% \small
% \begin{align} \label{eq:QAT}
% \hat{\theta} &= q \times \Big\lfloor{\text{clip}\Big(\frac{\theta}{q}, -Q_N, Q_P\Big)} \Big\rceil
% \end{align}
% \normalsize
% where $\text{clip}(y; r_1; r_2)$ returns $y$ with values below $r_1$ set to $r_1$ and values above $r_2$ set to $r_2$; and $\lfloor y \rceil$ rounds $y$ to the nearest integer; $Q_P$
% and $Q_N$ represent positive and negative quantization levels, respectively. For $b$-bit quantization, for activations we set $Q_N=0$ and $Q_P= 2^b-1$, and for weights we set $Q_N=2^{b-1}$ and $Q_P= 2^{b-1}-1$. 

In QAT, weights and activations are quantized, and  the forward and backward passes are performed over quantized weights and activations, while the original floating-point weights are updated \cite{lsq, rokh2023comprehensive}. A fundamental challenge in the QAT formulation stems from the lack of an interpretable gradient for the quantization function, making gradient-based training impractical without ad hoc approximation. One widely adopted approach to mitigate this challenge is, by employing the straight-through estimator (STE) \cite{bengio2013estimating, hinton2012ste}, to approximate the true gradient during training. Essentially, this involves approximating the gradient of the rounding operator as 1 within the quantization limits \cite{lsq,dorefa}. 

% The STE gradient approximation has gained widespread adoption in recent literature, effectively narrowing the accuracy gap between quantized and full-precision models across various tasks and network architectures \cite{Gupta2015, dorefa, krishnamoorthi, tqt, lsq, lsq+, whitepaper}.

Some studies showed that the derivative of the clipped ReLU provides a superior approximation compared to the vanilla STE (derivative of identity) \cite{guassian_quant, understanding_STE}. Notably, the former effectively sets the gradient outside the quantization grid to zero, and has become the predominant formulation in QAT literature \cite{pact2018, lsq, differentiablequantization}. In addition, much prior research concentrated on highlighting the sub-optimality of the STE as a gradient approximator, and  proposed some alternative gradient estimators which could be mainly categorized into two types: (i) multiplicative methods which apply a scaling to the gradient vectors \cite{EWGS,PBGS,DSQ,quantization_nets}, and (ii) additive methods that add an input-independent term to the gradient \cite{qat_tinyml,binreg,profit,metaquant}.

Nonetheless, while these approaches offer alternatives to STE, they all rely on certain approximations that can result in sub-optimal results. In addition, neither weight entropy nor activation entropy is considered. 

% \subsection{Accounting for Compression during Training} \label{sec:comp-aware}
% Some papers explicitly account for compression while training DNNs. For instance, \cite{alvarez2017compression} introduced a regularization term that encourages the parameter matrix of each layer to have low rank during training so that some parameters of the DNN can be pruned in the post-processing stage. Similarly, the works in \cite{aytekin2019compressibility,baskin2021cat} introduced other types of regularization terms to train compressible DNNs.

% In addition, the authors in \cite{chmiel2020robust} showed that uniform distribution of weights are more robust to quantization than normally distributed weights. To this end, they introduced KURE (KUrtosis REgularization) to minimize the kurtosis of weights and ensure a uniform distribution. \cite{han2021improving} proposed a method to constrain weights into predefined bins based on the quantization bit-width. However, selecting these bins can be challenging, and the model's performance is highly sensitive to the bin selection. \cite{kundu2024r} introduced range regularization, which imposes constraints on the weights to ensure they fall within predefined value ranges.

% However, in general, the methodologies proposed by these works lack analytic justifications and primarily  rely on empirical observations. More importantly, they do not satisfy all four criteria \textit{C1-C4} simultaneously as achieved by CDL. 

\subsection{Modes of Parallelism} \label{sec:mode-par}

Training large models poses significant challenges for two reasons: (i) it is no longer feasible to accommodate the parameters of these models in the main memory, even with the largest GPU configurations available (such as NVIDIA's 80GB-A100 cards); (ii) even if the model can be fit into a single GPU (e.g., by utilizing techniques like swapping parameters between host and device memory as in ZeRO-Offload \cite{ren2021zero}), the high volume of compute operations required can lead to unrealistically long training times (for instance, training GPT-3 with 175 billion parameters \cite{brown2020language} would necessitate approximately 288 years using a single V100 GPU \cite{narayanan2021efficient}). This underscores the need for parallelism.

Here, we provide a brief overview of parallelism techniques; specifically, we discuss two key approaches: model parallelism and data parallelism.

\subsubsection{Model parallelism} \label{sec:model-par}
Pipeline model parallelism divides the model into multiple chunks of layers and distributes them across different devices \cite{huang2019gpipe,kosson2021pipelined,yang2021pipemare}. During the forward pass, the activation outputs from one stage are sent to the next stage on a different device. During the backward pass, gradients are propagated back through the pipeline in a similar fashion.

Nonetheless, this approach incurs communication overhead between devices when sending activations between pipeline stages. This can potentially negate some of the speedup benefits, and can impact performance, particularly on systems with slower network connections.

\subsubsection{Data parallelism} \label{sec:data-par}
In data parallelism, the DNN  is replicated across multiple devices \cite{dean2012large,ben2019demystifying}. The training dataset is divided into mini-batches. Each device receives different mini-batches. Devices simultaneously process their assigned mini-batches, computing weight updates (or gradients) for the same model. Then, the devices update their local models by averaging the weights/gradients received from all the devices. 

To facilitate model/data parallelism, it is desirable for weights and activations to be compressible at any stage of training.

\section{Notation and Preliminaries}\label{sec:prelem}

\subsection{Notation} \label{sec:notation}
For a positive integer $N$,  let $[N]\triangleq \{1,\dots,N\}$. Scalars are denoted by lowercase letters (e.g., $w$), and vectors by bold-face  letters (e.g., $\boldsymbol{w}$). The $i$-th element of vector $\boldsymbol{w}$ is denoted by $\boldsymbol{w}[i]$. Also, the length of vector $\boldsymbol{w}$ is denoted by $|\boldsymbol{w}|$.

The probability of event $E$ is denoted by $P(E)$. In addition, for a discrete random variable $X$, its probability mass function, expected value, and variance are denoted by $\mathbb{P}_X$, $\mathbb{E}\{X\}$,  and $\text{Var}\{X\}$, respectively. For a $C$-dimensional probability distribution $P$, denote by $\mathsf{H}(P) = \sum_{c \in [C]} - P[c] \log P[c]$, the Shannon entropy of $P$. The softmax operation over the vector $\boldsymbol{w}$ is denoted by $\boldsymbol{\sigma}(\boldsymbol{w})$. That is, for any $i \in [|\boldsymbol{w}|]$
$$\boldsymbol{\sigma}(\boldsymbol{w})[i] = \frac{e^{\boldsymbol{w}[i]}}{\sum_{j \in [|\boldsymbol{w}|]} e^{\boldsymbol{w}[j]}}. $$

\subsection{Preliminaries}
In ordinary multi-class classification, let $\mathcal{X} \subseteq \mathbb{R}^{d}$ be the instance space, and $\mathcal{Y}=[C]$ be the label space, where $C$ is the number of classes.

A DNN is a cascade of $L$ layers. For both fully-connected and convolutional layers, denote by $\boldsymbol{w}_l$ the flattened vector representation of weight parameters at layer $l$. Hence the number of  weight parameters at layer $l$ equals $|\boldsymbol{w}_l |  $.  In addition, $\boldsymbol{w} = \{ \boldsymbol{w}_l \}_{l \in [L]}$ represents all the parameters in the DNN.

The input to the first layer of the DNN is a raw sample $\boldsymbol{x}_0 \in \mathcal{X}$. The output of the $l$-th layer, which is also called an activation map, is denoted by vector $\boldsymbol{x}_l$. Hence, the number of activations at the output of the $l$-th layer is equal to $|\boldsymbol{x}_l | $, and the output logit vector is represented by $\boldsymbol{x}_L$. Let  $\boldsymbol{x} = \{ \boldsymbol{x}_l \}_{l=1}^{L-1}$.

Let $(X, Y)$ be a pair of random variables, the distribution of which governs a training set, where $X \in \mathcal{X}  $ represents the raw input sample and $Y$ is the ground truth label of $X$. In traditional DL, one trains a classifier by solving the following minimization problem
\begin{equation} \label{eq:loss}
\min_{\boldsymbol{w} } \mathbb{E}_{(X,Y)} \{\mathcal{L} \left( X,Y, \boldsymbol{w} \right)\},  
\end{equation}
where $\mathcal{L} ( \boldsymbol{x}_0, y,  \boldsymbol{w}) $ is a real-valued loss function (typically the CE loss).

\section{Probabilistic Quantization Method} \label{sec:quant}
In this section, we first propose a probabilistic quantization method in Subsection \ref{sec:random} that forms the foundation for the CDL framework. Next, in Subsection \ref{sec:soft}, we introduce its \textit{soft deterministic} approximation. 

\begin{figure*}
\centering  \includegraphics[width=1\textwidth]{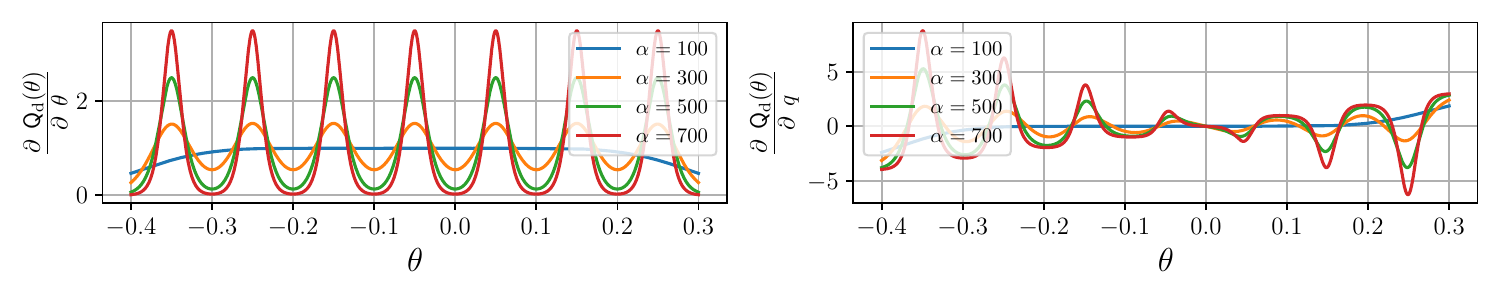}
\vskip -0.23in
  \caption{Illustration of the partial derivatives of $\mathsf{Q}_{\rm d}(\theta)$ w.r.t. $\theta$ (left), and $q$ (right) for $\alpha=\{ 100,300,500,700\}$, where $b$ and $q$ equal $3$ and $0.1$, respectively.} \label{fig:partial}
\vskip -0.1in  
\end{figure*}

\subsection{Probabilistic Quantizer $\mathsf{Q}_{\rm p}(\cdot)$} \label{sec:random}

Consider $b$-bit uniform quantization\footnote{The proposed probabilistic quantization method can be easily extended to non-uniform quantization (see Appendix \ref{app:non-uniform}).}. Let $\mathcal{A}$ denote its corresponding quantization index set 
\begin{align}
\mathcal{A}&= \{ -2^{b-1}, -2^{b-1}+1, \dots, 2^{b-1}-2,2^{b-1}-1 \}.   
\end{align}
For convenience, $\mathcal{A}$ is also regarded as a vector of dimension $2^b$. Multiplying $\mathcal{A}$ by a quantization step-size $q$, we get the corresponding reproduction alphabet
 \begin{align}
  \hat{\mathcal{A}} = q \times [-2^{b-1}, -2^{b-1}+1, \dots, 2^{b-1}-2,2^{b-1}-1].    
\end{align}
Again, we will regard $ \hat{\mathcal{A}} $ as both a vector and set.

Let $\theta$ represent either a weight parameter or an activation of a DNN. We want to quantize $\theta$ randomly to $\hat{\theta} \in  \hat{\mathcal{A}}$. To this end, we introduce a trainable CPMF $ P_{\alpha}(\cdot | \theta)  $ over the reproduction alphabet $ \hat{\mathcal{A}}  $ or equivalently the index set $ \mathcal{A} $ given $\theta$, where $\alpha >0 $ is a trainable parameter: 
 \begin{equation} \label{eq:CPMF}
     P_{\alpha}(\hat{\theta} | \theta) = \frac{e^{-\alpha (\theta - \hat{\theta})^2}}{\sum_{j \in \mathcal{A}} e^{-\alpha (\theta - jq)^2 }}, ~\forall \hat{\theta} = i q, ~i \in \mathcal{A}.
 \end{equation}
Define the following notation
\begin{align}
\big[\theta \big]_{2^b} =   [\underbrace{\theta, \dots, \theta}_{\text{$2^b$ times}}].
\end{align}
 Regard the CPMF $ P_{\alpha}(\cdot | \theta)  $ as a vector of dimension $2^b$. Then it can also be easily computed via the softmax operation:
\begin{align} \label{eq:PMF}
 \big[P_{\alpha}(\cdot | \theta) \big] &= \boldsymbol{\sigma} \Big(- \alpha \times \left(  \big[ \theta \big]_{2^b} -   \hat{\mathcal{A}} \right)^2 \Big),
\end{align}
where $\boldsymbol{\sigma}(\cdot)$ denotes the softmax operation. 

With the CPMF $ P_{\alpha}(\cdot | \theta)  $,  $\theta$ is now quantized to each  $\hat{\theta}   \in \hat{\mathcal{A}}  $ with probability $P_{\alpha}(\hat{\theta} | \theta)$. Denote this random mapping by 
\begin{align} \label{eq:Qprob}
\hat{\theta} =  \mathsf{Q}_{\rm p}(\theta).   
\end{align}
Note that given $\theta$, $\hat{\theta}  $ is a random variable taking values in $ \hat{\mathcal{A}}  $ with distribution $ P_{\alpha}(\cdot | \theta)  $.

\begin{remark}
Note that although $\mathsf{Q}_{\rm p}(\theta)  $ is a probabilistic quantizer, it quantizes $\theta$ into the nearest point in $  \hat{\mathcal{A}} $ with probability approaching $1 $ as $\alpha \to \infty$. Therefore, when $\alpha$ is large, $\mathsf{Q}_{\rm p}(\theta)  $ is equal to the uniformly quantized value with high probability. 
\end{remark}

When the probabilistic quantizer $\mathsf{Q}_{\rm p}(\theta)  $ is employed in DL, how would one compute the gradients w.r.t. $\theta$ and $q$ during backward passes? This is addressed in the next subsection.

\subsection{\textit{Soft Deterministic} Quantizer $\mathsf{Q}_{\rm d}(\cdot)$} \label{sec:soft}

Given $\theta$, let us first analyze the expected distortion (defined as the squared error) between $\theta$ and $\mathsf{Q}_{\rm p}(\theta)  $. To this end, consider the conditional expectation of $\hat{\theta}$ given $\theta$, i.e., 
\begin{align} \label{eq:softQ}
\mathsf{Q}_{\rm d}(\theta) = \mathbb{E} \{ \hat{\theta} |\theta \}  = \sum_{i \in \mathcal{A}} P_{\alpha}(iq|\theta) ~iq.    
\end{align}
We have the following proposition. 
\begin{proposition} \label{prop:distortion}
For any $\theta$ and $\alpha > 0$, 
\begin{align} \nonumber
\mathbb{E} \big\{ (\theta - \mathsf{Q}_{\rm p}(\theta) )^2 ~ | ~ \theta \big\} &= (\theta  - \mathsf{Q}_{\rm d}(\theta))^2 + \text{Var} \big\{ \mathsf{Q}_{\rm p}(\theta) ~|~\theta \big\},   
\end{align}  
where $ \text{Var} \big\{ \mathsf{Q}_{\rm p}(\theta) ~|~\theta \big\} $ is the conditional variance of $  \mathsf{Q}_{\rm p}(\theta)   $ given $\theta$. 
\end{proposition}

\begin{proof}
Please refer to Appendix \ref{app:dist} for the proof. 
\end{proof}

When $\alpha$ is sufficiently large, the conditional variance $\text{Var} \big\{ \mathsf{Q}_{\rm p}(\theta) ~|~\theta  \big\}$ becomes negligible. Thus, as per Proposition \ref{prop:distortion}, the expected distortion of the probabilistic quantizer is approximately equal to the distortion between $\theta$ and $ \mathsf{Q}_{\rm d}(\theta) $. Note that $ \mathsf{Q}_{\rm d}(\theta) $ is deterministic. This suggests that one may use $ \mathsf{Q}_{\rm d}(\theta) $ to approximate the probabilistic quantizer $  \mathsf{Q}_{\rm p}(\theta)   $. We refer to  $ \mathsf{Q}_{\rm d}(\cdot ) $ as a \textit{soft deterministic} quantizer or approximation.  The quantization is \textit{soft} in the sense that the output is not strictly in the reproduction alphabet $ \hat{\mathcal{A}}  $,  but rather maintains a degree of continuous representation. 

In view of \eqref{eq:softQ} and \eqref{eq:CPMF}, the \textit{soft deterministic} quantizer  $ \mathsf{Q}_{\rm d}(\cdot ) $ as a function of $\theta$ and $q$ is analytic and has nice analytic formulas for computing its partial derivatives w.r.t. $\theta$ and $q$, as shown in Lemma~\ref{lemma:partial} below, which is proved in Appendix \ref{app:partial}.

\begin{lemma} \label{lemma:partial}
The partial derivatives of $\mathsf{Q}_{\rm d}(\theta)$ w.r.t. $q$ and
$\theta$ are obtained as
\begin{subequations}\label{eq:partial}
\begin{align} \label{eq:partialW}
\frac{\partial \mathsf{Q}_{\rm d}(\theta)}{\partial \theta} &=  2\alpha \text{Var}  \big\{ \mathsf{Q}_{\rm p}(\theta) \big\}, \\
\frac{\partial \mathsf{Q}_{\rm d}(\theta)}{\partial q} &= \frac{1}{q} \Big( \mathbb{E} \big\{ \mathsf{Q}_{\rm p}(\theta) \big\} + (2\alpha  \theta) \text{Var}  \big\{ \mathsf{Q}_{\rm p}(\theta) \big\} \nonumber \\ \label{eq:partialQ}
& \qquad -(2\alpha ) \text{Skew}_{\rm u} \big\{ \mathsf{Q}_{\rm p}(\theta) \big\} \Big),
\end{align}
\end{subequations}
where for any random variable $X$,
  \[ 
\text{Skew}_{\rm u} (X) \defeq \sum_{x} x^3 \mathbb{P}_X(x) -\Big(\sum_{x} x \mathbb{P}_X(x) \Big) \Big( \sum_{x} x^2 \mathbb{P}_X(x)\Big). \]
\end{lemma}
To grasp a better understanding, $\frac{\partial \mathsf{Q}_{\rm d}(\theta)}{\partial \theta}$ and $\frac{\partial \mathsf{Q}_{\rm d}(\theta)}{\partial q}$ are depicted in Fig. \ref{fig:partial} for different values of $\alpha$. For each $\theta$, denote its nearest value in $\hat{\mathcal{A}}    $ by $\lfloor \theta \rceil_q$. It is informative to note from  Fig. \ref{fig:partial}  that as $\theta$ moves away from  $\lfloor \theta \rceil_q$ in either direction, $\frac{\partial \mathsf{Q}_{\rm d}(\theta)}{\partial \theta}$ increases. This important property bodes well with quantization and helps the training process push $\theta$ towards points inside $\hat{\mathcal{A}}$, which is more or less confirmed by Fig. \ref{fig:distweight} in Section~\ref{sec:exp}.

\begin{figure*}
\centering  \includegraphics[width=1\textwidth]{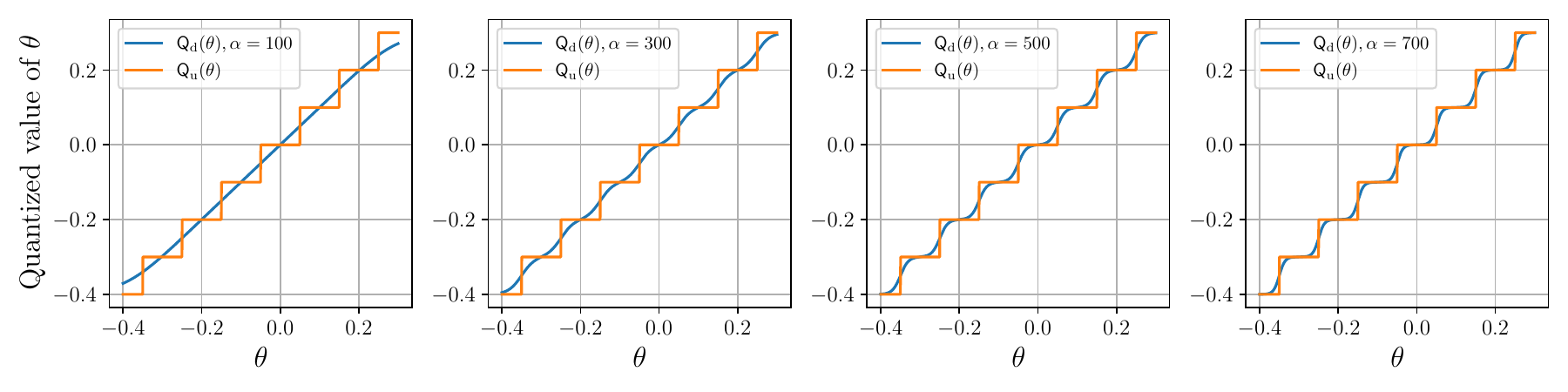}
  \vskip -0.23in
  \caption{Illustration of  $\mathsf{Q}_{\rm u}(\cdot)$ vs $\mathsf{Q}_{\rm d}(\cdot)$ with $\alpha=\{ 100,300,500,700\}$, where $b$ and $q$ are set to $3$ and $0.1$, respectively.} \label{fig:softVsU}
  \vskip -0.2in
\end{figure*}

We conclude this section by comparing the \textit{soft deterministic} quantizer $\mathsf{Q}_{\rm d}(\theta)$ with the real uniform quantizer $\mathsf{Q}_{\rm u}(\theta) = \lfloor \theta \rceil_q$. 
Fig. \ref{fig:softVsU} depicts the curves of $\mathsf{Q}_{\rm u}(\theta)$ and $\mathsf{Q}_{\rm d}(\theta)$ for different $\alpha$ values. As observed, as $\alpha$ increases, $\mathsf{Q}_{\rm d}(\cdot)$  converges to the uniform quantizer.

% \begin{remark}
% In standard quantization methods, inter alia round-to-nearest integer, the parameter $\theta$ is deterministically mapped to an element of $\bm{\theta}$. We note that this is also the case for stochastic quantization methods, although a probabilistic approach is used, the quantized value is still chosen from $\bm{\theta}$.    
% \end{remark}

\section{CDL and R-CDL} \label{sec:CDL}
\begin{figure*}
\centering  \includegraphics[width=1\textwidth]{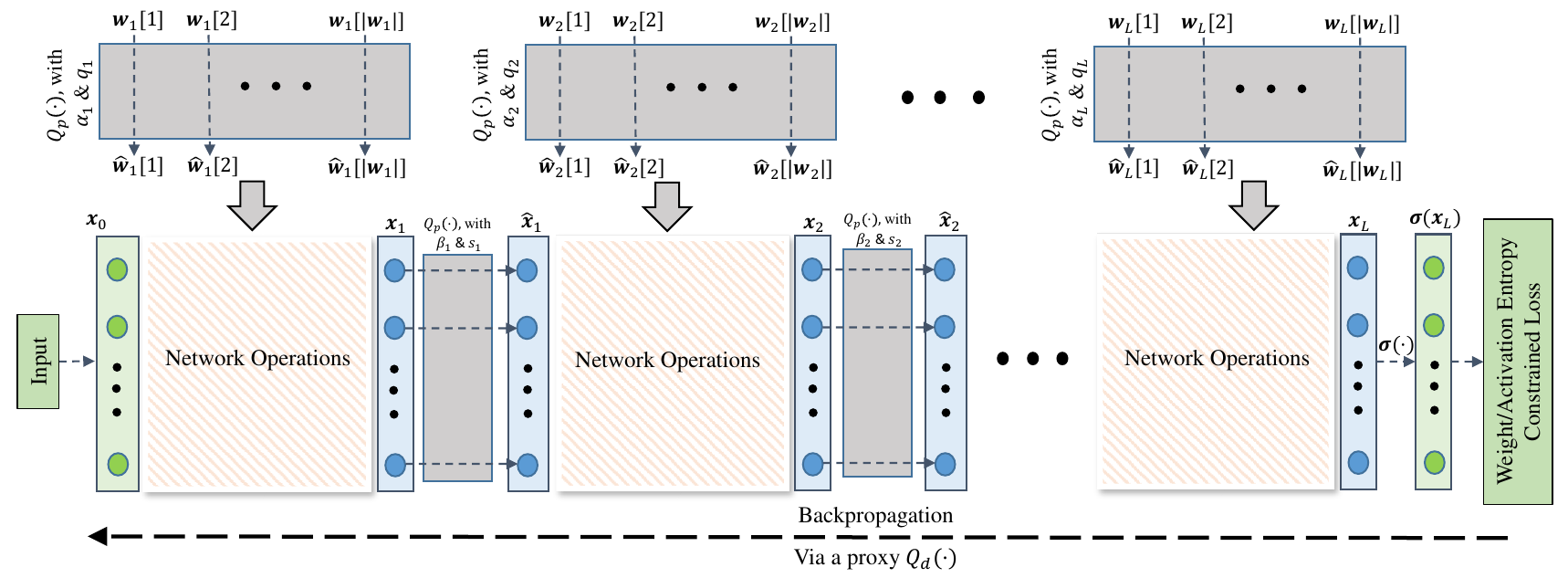}
  \vskip -0.1in
  \caption{Illustration of the CDL's mechanism.} \label{fig:CDL}
  \vskip -0.1in
\end{figure*}

This section is devoted to the detailed description of the inner workings of CDL and R-CDL. As shown in Fig.~\ref{fig:CDL}, key elements of CDL include: (i) incorporating trainable probabilistic quantizers $\mathsf{Q}_{\rm p}(\cdot)$ to quantize both model weights and activations; (ii) enforcing entropy constraints on quantized weights and activations during training, and (iii) using the \textit{soft deterministic} quantizers $\mathsf{Q}_{\rm d}(\cdot)$
corresponding to  $\mathsf{Q}_{\rm p}(\cdot)$  as a proxy to facilitate gradient computation in backpropagation.
To facilitate our discussion and make it easier to understand, we decompose it into several parts: (i) incorporation of trainable probabilistic quantizers $\mathsf{Q}_{\rm p}(\cdot)$ into CDL; (ii) computation of gradients of the loss function without entropy constraint via backpropagation deploying $\mathsf{Q}_{\rm d}(\cdot)$;  (iii) entropy computation of quantized weights and activations; (iv) new objective function with entropy constraints; (v) new overall learning algorithm; and (vi) R-CDL with trainable quantizers $\mathsf{Q}_{\rm d}(\cdot)$.

\subsection{Incorporating Trainable Probabilistic Quantizers $\mathsf{Q}_{\rm p}(\cdot)$ Into CDL} \label{sec:CDLw}
% In this section, we discuss how (\textit{soft}) probabilistic quantization method is employed in CDL framework. 

Consider $b$-bit precision quantization. In CDL, both weights and activations are quantized using trainable probabilistic quantizers $\mathsf{Q}_{\rm p}(\cdot)$ with the following specifications.

First, due to the adoption of ReLU function in DNN architecture, activations take only non-negative values. As such, we use the following sets $ \mathcal{A}$ and $ \mathcal{B} $ as the quantization index sets for weights and activations, respectively 
\begin{subequations}\label{eq:repo_alphabet}
\begin{align}
\mathcal{A} &= \{ -2^{b-1}, -2^{b-1}+1, \dots, 2^{b-1}-2,2^{b-1}-1 \}, \\
\mathcal{B} &= \{0, 1, \dots, 2^{b}-2,2^{b}-1 \}.
\end{align}
\end{subequations}

Second, to accommodate different characteristics of different layers, we use different trainable probabilistic quantizers $\mathsf{Q}_{\rm p}(\cdot)$ for different layers. Let  $q_l$ and $s_l$ denote the quantization step-sizes for weights and activations at layer $l$, $l \in [L]$, respectively. Then the corresponding reproduction alphabets for weights $\boldsymbol{w}_l$ and activations $\boldsymbol{x}_l$ at layer $l$, become
% \small
\begin{subequations}\label{eq:alphabet}
\begin{align}
\hat{\mathcal{A}}_l &= q_l \times [-2^{b-1}, -2^{b-1}+1, \dots, 2^{b-1}-2,2^{b-1}-1], \\
\hat{\mathcal{B}}_l &= s_l \times [0, 1, \dots, 2^{b}-2,2^{b}-1 ].
\end{align}
\end{subequations}
% \normalsize

Finally,  as per \eqref{eq:PMF}, we deploy (i) a probabilistic quantizer $\mathsf{Q}_{\rm p}(\cdot)$ with trainable parameters $q_l>0 $ and $\alpha_l >0 $ to quantize each entry of $\boldsymbol{w}_l$ according to the following CPMF
% \small
\begin{align} \label{eq:PMFw}
&\big[P_{\alpha_l}( \cdot  | \boldsymbol{w}_l[i])\big] = \boldsymbol{\sigma} \Big(- \alpha_l \times \big(  \big[\boldsymbol{w}_l[i]\big]_{2^b} -\hat{\mathcal{A}}_l \big)^2 \Big),
\end{align}
and (ii) another probabilistic quantizer $\mathsf{Q}_{\rm p}(\cdot)$ with trainable parameters $s_l >0 $ and $\beta_l > 0$ to quantize
each entry of $\boldsymbol{x}_l$ according to the following CPMF
\begin{align}  \label{eq:PMFa}
&\big[P_{\beta_l}( \cdot | \boldsymbol{x}_l[j])\big] = \boldsymbol{\sigma} \Big(- \beta_l \times \big(  \big[\boldsymbol{x}_l[j]\big]_{2^b} - \hat{\mathcal{B}}_l \big)^2 \Big). 
\end{align}

\begin{remark}
Note that the activations at the last layer are not required to be quantized. As such, trainable quantization parameters are $\boldsymbol{q} = \{ q_l \}_{l \in [L]}$, $\boldsymbol{s} = \{ s_l \}_{l \in [L-1]}$, $\boldsymbol{\alpha} = \{ \alpha_l \}_{l \in [L]}$, and $\boldsymbol{\beta} = \{ \beta_l \}_{l \in [L-1]}$, which will be learned jointly with weights $ \boldsymbol{w} $  during the training process. 
\end{remark}

Henceforth, the training calculation (forward and backward) passes are performed over these quantized weights and activations. Accordingly, with the incorporation of probabilistic quantizers, the conventional loss in DL becomes
\begin{align} \label{eq:Qloss}
\mathcal{L} \Big( \boldsymbol{x}_0, y, \{\mathsf{Q}_{\rm p}(\boldsymbol{w}), \mathsf{Q}_{\rm p}(\boldsymbol{x}) \}\Big),    
\end{align}
where the notation $\{\mathsf{Q}_{\rm p}(\boldsymbol{w}), \mathsf{Q}_{\rm p}(\boldsymbol{x}) \}$ implies that both weights and activations are quantized according to $\mathsf{Q}_{\rm p}(\cdot)$ operation.

\subsection{Computation of Gradients of the Loss Function \eqref{eq:Qloss}} \label{sec:CDLgrad}

The loss function \eqref{eq:Qloss} involves random jumps from $ \boldsymbol{w}_l[i] $ to $ \mathsf{Q}_{\rm p}(\boldsymbol{w}_l[i])  $ and from $ \boldsymbol{x}_l[j] $ to $ \mathsf{Q}_{\rm p}(\boldsymbol{x}_l[j])  $  and hence is not differentiable. How to compute its gradients with respect to $ \boldsymbol{w}$, $ \boldsymbol{x} $, $  \boldsymbol{q}  $, $ \boldsymbol{s} $, $  \boldsymbol{\alpha} $, and $ \boldsymbol{\beta} $? To overcome this difficulty, we invoke Proposition~\ref{prop:distortion} and Lemma~\ref{lemma:partial}, and  use $\mathsf{Q}_{\rm d}(\cdot)$  as a proxy for gradient calculation as explained below.

Firstly, following \eqref{eq:softQ}, the quantized values of $\boldsymbol{w}_l[i]$ and $\boldsymbol{x}_l[j]$ by the \textit{soft deterministic} quantizers $\mathsf{Q}_{\rm d}(\cdot)$  are 
% \small
\begin{subequations}\label{eq:softboth}
\begin{align}
\mathsf{Q}_{\rm d}(\boldsymbol{w}_l[i]) &= \mathbb{E} \{ \hat{w} |\boldsymbol{w}_l[i] \} = \sum_{j \in \mathcal{A}} P_{\alpha_l}( j q_l | \boldsymbol{w}_l[i]) ~j q_l, \\
\mathsf{Q}_{\rm d}(\boldsymbol{x}_l[j]) &= \mathbb{E} \{  \hat{x}|\boldsymbol{x}_l[j] \} = \sum_{i\in \mathcal{B}} P_{\beta_l}(i s_l | \boldsymbol{x}_l[j]) ~i s_l.
\end{align}
\end{subequations}
% \normalsize
Next, we compute the partial derivatives of $\mathcal{L} ( \boldsymbol{x}_0, y,  \{\mathsf{Q}_{\rm p}(\boldsymbol{w}), \mathsf{Q}_{\rm p}(\boldsymbol{x}) \} )$ w.r.t. $\boldsymbol{w}_l[i]$. To this end, we first need to find the partial derivatives w.r.t. $\boldsymbol{x}_l[j]$ as follows
\begin{align}\label{eq:partialX1}
&\frac{\partial \mathcal{L} \Big( \boldsymbol{x}_0,y, \{\mathsf{Q}_{\rm p}(\boldsymbol{w}), \mathsf{Q}_{\rm p}(\boldsymbol{x}) \}\Big)}{\partial \boldsymbol{x}_l[j]} \\ \label{eq:partialX2}
&= \frac{\partial \mathcal{L} \Big( \boldsymbol{x}_0,y,  \{\mathsf{Q}_{\rm p}(\boldsymbol{w}), \mathsf{Q}_{\rm p}(\boldsymbol{x}) \}\Big)}{\partial \mathsf{Q}_{\rm p}(\boldsymbol{x}_l[j])} \frac{\partial  \mathsf{Q}_{\rm p}(\boldsymbol{x}_l[j]) }{\partial \boldsymbol{x}_l[j]} \\ \label{eq:partialX3}
& \approx \frac{\partial \mathcal{L} \Big( \boldsymbol{x}_0,y,  \{\mathsf{Q}_{\rm p}(\boldsymbol{w}), \mathsf{Q}_{\rm p}(\boldsymbol{x}) \}\Big)}{\partial \mathsf{Q}_{\rm p}(\boldsymbol{x}_l[j])} \frac{\partial  \mathsf{Q}_{\rm d}(\boldsymbol{x}_l[j]) }{\partial \boldsymbol{x}_l[j]},
\end{align}
where the approximation in \eqref{eq:partialX3} is obtained by replacing $\mathsf{Q}_{\rm p}(\boldsymbol{x}_l[j])$ in the second partial derivative of \eqref{eq:partialX2} with its differentiable approximation $\mathsf{Q}_{\rm d}(\boldsymbol{x}_l[j])$. The first partial derivative in \eqref{eq:partialX3} can be calculated by backpropagation over the layers of DNN, and the second partial derivative could be computed according to  \eqref{eq:partialW} with corresponding parameters. 
Now, the partial derivatives w.r.t.  $\boldsymbol{w}_l[i]$ can be calculated as
% \small
\begin{align}\label{eq:partialw} 
&\frac{\partial \mathcal{L} \Big( \boldsymbol{x}_0,y,   \{\mathsf{Q}_{\rm p}(\boldsymbol{w}), \mathsf{Q}_{\rm p}(\boldsymbol{x}) \}\Big)}{\partial \boldsymbol{w}_l[i]} \\  \nonumber 
&= \sum_{j }
\frac{\partial \mathcal{L} \Big( \boldsymbol{x}_0,y,  \{\mathsf{Q}_{\rm p}(\boldsymbol{w}), \mathsf{Q}_{\rm p}(\boldsymbol{x}) \}\Big)}{\partial \boldsymbol{x}_l[j]} \frac{\partial \boldsymbol{x}_l[j]}{\partial \mathsf{Q}_{\rm p}(\boldsymbol{w}_l[i])} \frac{\partial  \mathsf{Q}_{\rm p}(\boldsymbol{w}_l[i]) }{\partial \boldsymbol{w}_l[i]} \\ \label{eq:partialW3}
& \approx  \frac{\partial  \mathsf{Q}_{\rm d}(\boldsymbol{w}_l[i]) }{\partial \boldsymbol{w}_l[i]} \sum_{j }
\frac{\partial \mathcal{L} \Big(\boldsymbol{x}_0,y,  \{\mathsf{Q}_{\rm p}(\boldsymbol{w}), \mathsf{Q}_{\rm p}(\boldsymbol{x}) \}\Big)}{\partial \boldsymbol{x}_l[j]} \frac{\partial \boldsymbol{x}_l[j]}{\partial \mathsf{Q}_{\rm p}(\boldsymbol{w}_l[i])},
\end{align}
% \normalsize
where the summation above is taken over all $j$ corresponding to which $ \boldsymbol{x}_l[j] $ depends on $ \mathsf{Q}_{\rm p}(\boldsymbol{w}_l[i])  $.

In \eqref{eq:partialW3}, the first partial derivative is obtained using \eqref{eq:partialW} with corresponding parameters, and  the rest partial derivatives can be  obtained via backpropagation and  \eqref{eq:partialX3}.  

In addition, the partial derivative of $\mathcal{L} ( \boldsymbol{x}_0,y,  \{\mathsf{Q}_{\rm p}(\boldsymbol{w}), \mathsf{Q}_{\rm p}(\boldsymbol{x}) \} )$ w.r.t. $q_l$ is calculated as follows
\begin{align}
&\frac{\partial \mathcal{L} \Big( \boldsymbol{x}_0,y,  \{\mathsf{Q}_{\rm p}(\boldsymbol{w}), \mathsf{Q}_{\rm p}(\boldsymbol{x}) \}\Big)}{\partial q_l} \nonumber \\ \label{eq:Jgradqw}
& \approx \sum_{i \in [|\boldsymbol{w}_l |] }  \frac{\partial \mathcal{L} \Big( \boldsymbol{x}_0,y,   \{\mathsf{Q}_{\rm p}(\boldsymbol{w}), \mathsf{Q}_{\rm p}(\boldsymbol{x}) \}\Big)}{\partial \mathsf{Q}_{\rm p}(\boldsymbol{w}_l[i])} \frac{\partial  \mathsf{Q}_{\rm d}(\boldsymbol{w}_l[i]) }{\partial q_l}, 
\end{align}
where the second partial derivative in \eqref{eq:Jgradqw} could be computed according to \eqref{eq:partialQ}. Similarly,  
\begin{align}
&\frac{\partial \mathcal{L} \Big( \boldsymbol{x}_0,y,   \{\mathsf{Q}_{\rm p}(\boldsymbol{w}), \mathsf{Q}_{\rm p}(\boldsymbol{x}) \}\Big)}{\partial s_l} \nonumber \\ \label{eq:Jgradqx}
& \approx \sum_{j \in [|\boldsymbol{x}_l|] }  \frac{\partial \mathcal{L} \Big(\boldsymbol{x}_0,y,   \{\mathsf{Q}_{\rm p}(\boldsymbol{w}), \mathsf{Q}_{\rm p}(\boldsymbol{x}) \}\Big)}{\partial \mathsf{Q}_{\rm p}(\boldsymbol{x}_l[j])} \frac{\partial  \mathsf{Q}_{\rm d}(\boldsymbol{x}_l[j]) }{\partial s_l}.
\end{align}
Also, for the mapping parameters $\alpha_l$  and $\beta_l$, we use the following approximations
\begin{align}
&\frac{\partial \mathcal{L} \Big( \boldsymbol{x}_0,y,  \{\mathsf{Q}_{\rm p}(\boldsymbol{w}), \mathsf{Q}_{\rm p}(\boldsymbol{x}) \}\Big)}{\partial \alpha_l} \nonumber \\ \label{eq:Jgradq_alpha_w}
& \approx \sum_{i \in [|\boldsymbol{w}_l|]}  \frac{\partial \mathcal{L} \Big( \boldsymbol{x}_0,y,  \{\mathsf{Q}_{\rm p}(\boldsymbol{w}), \mathsf{Q}_{\rm p}(\boldsymbol{x}) \}\Big)}{\partial \mathsf{Q}_{\rm p}(\boldsymbol{w}_l[i])} \frac{\partial  \mathsf{Q}_{\rm d}(\boldsymbol{w}_l[i]) }{\partial \alpha_l}, 
\end{align}
and 
\begin{align}
&\frac{\partial \mathcal{L} \Big( \boldsymbol{x}_0,y,  \{\mathsf{Q}_{\rm p}(\boldsymbol{w}), \mathsf{Q}_{\rm p}(\boldsymbol{x}) \}\Big)}{\partial \beta_l} \nonumber \\ \label{eq:Jgrad_alpha_q}
& \approx \sum_{j \in [|\boldsymbol{x}_l| ]}  \frac{\partial \mathcal{L} \Big( \boldsymbol{x}_0,y,  \{\mathsf{Q}_{\rm p}(\boldsymbol{w}), \mathsf{Q}_{\rm p}(\boldsymbol{x}) \}\Big)}{\partial \mathsf{Q}_{\rm p}(\boldsymbol{x}_l[j])} \frac{\partial  \mathsf{Q}_{\rm d}(\boldsymbol{x}_l[j]) }{\partial \beta_l} ,
\end{align}
where $  \frac{\partial  \mathsf{Q}_{\rm d}(\boldsymbol{w}_l[i]) }{\partial \alpha_l}  $ and $  \frac{\partial  \mathsf{Q}_{\rm d}(\boldsymbol{x}_l[j]) }{\partial \beta_l}   $ can be computed analytically.

\subsection{Entropy Computation of Quantized Weights and Activations} \label{sec:entropy}

Consider layer $l$. All weights at layer $l$ are randomly quantized according to the CPMF defined in \eqref{eq:PMFw}. Pick a weight $W$ randomly at layer $l$. Let $ \hat{W}$ denote its corresponding randomly quantized weight according to \eqref{eq:PMFw}. The marginal distribution of $W$ is the empirical distribution of weights at layer $l$. The MPMF of $\hat{W}$
is then given by 
\begin{align} \label{eq:approx1/n}
P_{\alpha_l}(\hat{w}) 
& = \frac{1}{|\boldsymbol{w}_l|} \sum_{i \in [|\boldsymbol{w}_l|]} P_{\alpha_l}(\hat{w} | \boldsymbol{w}_l[i]), ~\forall \hat{w} \in \hat{\mathcal{A}}_l .
\end{align}

If entropy coding (such as Huffman coding) is applied to encode all quantized weights at layer $l$, 
 the average number of bits per weight would be roughly equal to the Shannon entropy of $\hat{W}$
  \[ H(\hat{W}) = \mathsf{H}\big(\frac{1}{|\boldsymbol{w}_l|} \sum_{i \in [|\boldsymbol{w}_l|]} [ P_{\alpha_l}(\cdot | \boldsymbol{w}_l[i]) ] \big). \]
  Accordingly, the total number of bits required to represent all quantized weights at layer $l$ can be represented by 
\begin{align}
\mathsf{H}\big(\boldsymbol{w}_l\big) = |\boldsymbol{w}_l| ~\mathsf{H} \big( \frac{1}{|\boldsymbol{w}_l|} \sum_{i \in [|\boldsymbol{w}_l|]} [ P_{\alpha_l}( \cdot | \boldsymbol{w}_l[i]) ] \big),    
\end{align}
and the total number of bits required to represent all (quantized) weights of the whole DNN is 
\begin{align} \label{eq:totbit}
\mathsf{H}\big(\boldsymbol{w}\big) = \sum_{l \in [L]}  \mathsf{H}\big(\boldsymbol{w}_l\big), 
\end{align}
which can be referred to as the description complexity of the DNN model.

Similarly,  the total number of bits required to represent $\boldsymbol{x}_l$ is 
\begin{align}
\mathsf{H}\big(\boldsymbol{x}_l\big) = |\boldsymbol{x}_l|~ \mathsf{H} \big( \frac{1}{|\boldsymbol{x}_l|} \sum_{i \in [|\boldsymbol{x}_l|]} [ P_{\beta_l}(\cdot | \boldsymbol{x}_l[i]) ] \big),    
\end{align}
where $P_{\beta_l}(\cdot | \boldsymbol{x}_l[i]))$ is calculated according to \eqref{eq:PMFa}. And the total number of bits required to represent all quantized actvations of the whole DNN is
\begin{align}
\mathsf{H}\big(\boldsymbol{x}\big) = \sum_{l \in [L-1]}  \mathsf{H}\big(\boldsymbol{x}_l\big). 
\end{align}

\subsection{New Objective Function with Entropy Constraints} \label{sec:objective}

To make both weights and activations compressible at any stage of training, we now impose constraints on $\mathsf{H}\big(\boldsymbol{w}\big)$ and $\mathsf{H}\big(\boldsymbol{x}\big)$ during training. This can be effectively achieved by incorporating $\mathsf{H}\big(\boldsymbol{w}\big)$ and $\mathsf{H}\big(\boldsymbol{x}\big)$ into the objective function so that  $\mathsf{H}\big(\boldsymbol{w}\big)$ and $\mathsf{H}\big(\boldsymbol{x}\big)$ can be minimized jointly with
the original loss $\mathcal{L} ( \boldsymbol{x}_0,y,  \{\mathsf{Q}_{\rm p}(\boldsymbol{w}), \mathsf{Q}_{\rm p}(\boldsymbol{x})\} ) $. Thus the new objective function in CDL is
 \begin{align} 
 & \hat{\mathcal{L}} \Big( \boldsymbol{x}_0,y, \{\mathsf{Q}_{\rm p}(\boldsymbol{w}), \mathsf{Q}_{\rm p}(\boldsymbol{x})\}, 
 \boldsymbol{q}, \boldsymbol{s}, \boldsymbol{\alpha}, \boldsymbol{\beta} \Big)  \nonumber \\
  & \;   =   \mathcal{L} \Big( \boldsymbol{x}_0,y,  \{\mathsf{Q}_{\rm p}(\boldsymbol{w}), \mathsf{Q}_{\rm p}(\boldsymbol{x}) \} \Big)
   + \gamma \mathsf{H}(\boldsymbol{x})   + \lambda \mathsf{H}(\boldsymbol{w})  \label{eq:obj} \\
& \; =  \mathcal{L}_a \Big( \boldsymbol{x}_0,y,  \{\mathsf{Q}_{\rm p}(\boldsymbol{w}), \mathsf{Q}_{\rm p}(\boldsymbol{x}) \} \Big)  + \lambda \mathsf{H}(\boldsymbol{w}), \label{eq:obj2}
 \end{align}
where $\lambda \geq 0$ and $\gamma \geq 0$ are hyperparameters representing the trade-offs among the three terms in \eqref{eq:obj}, and
  \begin{align}
&  \mathcal{L}_a \Big( \boldsymbol{x}_0,y,  \{\mathsf{Q}_{\rm p}(\boldsymbol{w}), \mathsf{Q}_{\rm p}(\boldsymbol{x}) \} \Big) \nonumber \\ 
& \; = \mathcal{L} \Big( \boldsymbol{x}_0,y,  \{\mathsf{Q}_{\rm p}(\boldsymbol{w}), \mathsf{Q}_{\rm p}(\boldsymbol{x}) \} \Big)
   + \gamma \mathsf{H}(\boldsymbol{x}).  \label{eq:obj3}
  \end{align}

In CDL, the learning process is then to solve the following minimization problem
\begin{align}
& \min_{(\boldsymbol{q}, \boldsymbol{s}, \boldsymbol{\alpha}, \boldsymbol{\beta}, \boldsymbol{w} )}  
\mathbb{E} \Big\{ \hat{\mathcal{L}} \Big( X, Y,  \{\mathsf{Q}_{\rm p}(\boldsymbol{w}), \mathsf{Q}_{\rm p}(\boldsymbol{x})\}, 
 \boldsymbol{q}, \boldsymbol{s}, \boldsymbol{\alpha}, \boldsymbol{\beta} \Big) \Big \} \nonumber \\
&  = \min_{(\boldsymbol{q}, \boldsymbol{s}, \boldsymbol{\alpha}, \boldsymbol{\beta}, \boldsymbol{w})}
\Big\{ \mathbb{E} \Big\{ \mathcal{L}_a \big( X,Y, \{\mathsf{Q}_{\rm p}(\boldsymbol{w}), \mathsf{Q}_{\rm p}(\boldsymbol{x}) \}\big) \Big \}  + \lambda \mathsf{H}(\boldsymbol{w})  \Big\}   
 \label{eq:opt1} \\ 
& = \min_{(\boldsymbol{q}, \boldsymbol{s}, \boldsymbol{\alpha}, \boldsymbol{\beta},\boldsymbol{w} )}  
\Big\{ \lambda \mathsf{H}(\boldsymbol{w})  \nonumber \\ 
&  \qquad \qquad + \mathbb{E}_{X, Y, \mathsf{Q}_{\rm p}} \big\{ \mathcal{L}_a \big( X,Y, \{\mathsf{Q}_{\rm p}(\boldsymbol{w}), \mathsf{Q}_{\rm p}(\boldsymbol{x}) \}\big)   \big \} \Big\}.    \label{eq:opt2} 
\end{align}
In the above, \eqref{eq:opt1} is due to the fact that $ \mathsf{H}(\boldsymbol{w}) $ is deterministic given $ \boldsymbol{w} $; the expectation in \eqref{eq:opt2} is with respect to the random sample $(X, Y)$ and random quantizers $ \mathsf{Q}_{\rm p} $. 

In practice, when the joint distribution of the random sample $(X, Y)$ is unknown, the expectation in \eqref{eq:opt2} can be approximated by the corresponding sample mean over a mini-batch $ \mathcal{B} $ during learning, where for each sample instance $( \boldsymbol{x}_0, y) \in \mathcal{B}$, only one instance of each probabilistic quantizer $ \mathsf{Q}_{\rm p} $ is taken. In this case, the objective function in \eqref{eq:opt2} becomes
\begin{align}
& \mathcal{J}_{\mathcal{B}} \Big(  \{\mathsf{Q}_{\rm p}(\boldsymbol{w}), \mathsf{Q}_{\rm p}(\boldsymbol{x})\}, 
 \boldsymbol{q}, \boldsymbol{s}, \boldsymbol{\alpha}, \boldsymbol{\beta} \Big) = \lambda \mathsf{H}(\boldsymbol{w}) 
 \nonumber \\
& \;   + \frac{1}{|\mathcal{B}|} \sum_{ (\boldsymbol{x}_0, y) \in \mathcal{B}}
\mathcal{L}_a \Big( \boldsymbol{x}_0,y,  \{\mathsf{Q}_{\rm p}(\boldsymbol{w}), \mathsf{Q}_{\rm p}(\boldsymbol{x}) \} \Big),
      \label{eq:opt3}
\end{align}
where $  |\mathcal{B}| $ denotes the size of $ \mathcal{B}  $. The learning process is to iteratively solve the following minimization problem through a sequence of mini-batches
\small
\begin{align}
& \min_{(\boldsymbol{q}, \boldsymbol{s}, \boldsymbol{\alpha}, \boldsymbol{\beta}, \boldsymbol{w})}  
\Big \{ \lambda \mathsf{H}(\boldsymbol{w})  +  \frac{1}{|\mathcal{B}|} \sum_{ (\boldsymbol{x}_0, y) \in \mathcal{B}}
 \mathcal{L}_a \Big( \boldsymbol{x}_0,y,  \{\mathsf{Q}_{\rm p}(\boldsymbol{w}), \mathsf{Q}_{\rm p}(\boldsymbol{x}) \} \Big)
     \Big \}. \label{eq:joint_opt}
\end{align}
\normalsize

Let $\boldsymbol{\Omega} = (\boldsymbol{q}, \boldsymbol{s}, \boldsymbol{\alpha}, \boldsymbol{\beta}, \boldsymbol{w})$. Before we conclude this subsection, let us compute the gradient of $ \mathcal{J}_{\mathcal{B}} (  \{\mathsf{Q}_{\rm p}(\boldsymbol{w}), \mathsf{Q}_{\rm p}(\boldsymbol{x})\}, 
 \boldsymbol{q}, \boldsymbol{s}, \boldsymbol{\alpha}, \boldsymbol{\beta} )  $ with respect to $\boldsymbol{\Omega}$. 
 First note that the partial derivatives of  $\mathsf{H}(\boldsymbol{w})$  with respect to $(\boldsymbol{w}, \boldsymbol{q},  \boldsymbol{\alpha})  $   can be computed analytically. 

 Second, the gradient of $ \mathcal{L}_a ( \boldsymbol{x}_0,y,  \{\mathsf{Q}_{\rm p}(\boldsymbol{w}), \mathsf{Q}_{\rm p}(\boldsymbol{x}) \} )  $ with respect to $\boldsymbol{\Omega}$ can be computed via Eqs. \eqref{eq:partialX1} to \eqref{eq:Jgrad_alpha_q} and backpropagation with $\mathcal{L}$ replaced by $\mathcal{L}_a$ in all equations therein. Finally, we have 
 \begin{align}
& \frac{\partial \mathcal{J}_{\mathcal{B}} \Big(  \{\mathsf{Q}_{\rm p}(\boldsymbol{w}), \mathsf{Q}_{\rm p}(\boldsymbol{x})\}, 
 \boldsymbol{q}, \boldsymbol{s}, \boldsymbol{\alpha}, \boldsymbol{\beta} \Big)}{\partial \boldsymbol{\Omega}} = \lambda \frac{\partial \mathsf{H}(\boldsymbol{w}) }{\partial \boldsymbol{\Omega}}
 \nonumber \\
& \;   + \frac{1}{|\mathcal{B}|} \sum_{ (\boldsymbol{x}_0, y) \in \mathcal{B}}
\frac{\partial \mathcal{L}_a \Big( \boldsymbol{x}_0,y,  \{\mathsf{Q}_{\rm p}(\boldsymbol{w}), \mathsf{Q}_{\rm p}(\boldsymbol{x}) \} \Big)}{\partial \boldsymbol{\Omega}}. 
      \label{eq:opt4}
\end{align}

\subsection{Algorithm for Solving the Optimization in \eqref{eq:joint_opt}} \label{subsec:alg}
In order to solve the optimization problem in \eqref{eq:joint_opt} over mini-batches $\{\mathcal{B}_b\}_{b \in [B]}$, we deploy the conventional gradient descent (GD) strategy to update the five sets of parameters $\boldsymbol{q}, \boldsymbol{s}, \boldsymbol{\alpha}, \boldsymbol{\beta}$ and $\boldsymbol{w}$ in \eqref{eq:joint_opt}. The partial derivatives of $\mathcal{J}_{\mathcal{B}}(\cdot)$ w.r.t. these five sets of parameters can be computed via \eqref{eq:opt4}. 

\begin{algorithm}[!t]
\caption{CDL} 
\begin{algorithmic}[1] \label{alg:01}
\renewcommand{\algorithmicrequire}{\textbf{Input:}}
\renewcommand{\algorithmicensure}{\textbf{Output:}}
\REQUIRE All mini-batches $\{\mathcal{B}_b\}_{b \in [B]}$, maximum number of epochs $E_{\rm max}$, $\lambda$, $\gamma$, the quantization index sets $\mathcal{A}$ and $\mathcal{B}$, and learning rates $\{\eta_{w}, \eta_{q}, \eta_{s}, \eta_{\alpha},
\eta_{\beta}\}$. 

\STATE \textbf{Initialization:} Initialize  $\boldsymbol{\Omega}^0 = (\boldsymbol{q}^0, \boldsymbol{s}^0, \boldsymbol{\alpha}^0, \boldsymbol{\beta}^0, \boldsymbol{w}^0) $.

\FOR {$e=1$ to $E_{\rm max}$}
\FOR {$b=1$ to $B$}

\STATE Randomly quantize the weights $(\boldsymbol{w})^e_{b-1}$ according to $\mathsf{Q}_{\rm p}(\cdot)$ with parameters $(\boldsymbol{q})^e_{b-1}$ and $(\boldsymbol{\alpha})^e_{b-1}$.

\STATE For each sample $( \boldsymbol{x}_0, y) \in \mathcal{B}_b$, perform forward pass based on quantized weights  $\mathsf{Q}_{\rm p}( (\boldsymbol{w})^e_{b-1}  )$ while successively quantizing the activations according to $\mathsf{Q}_{\rm p}(\cdot)$ with parameters $(\boldsymbol{s})^e_{b-1}$ and $(\boldsymbol{\beta})^e_{b-1}$. 

\STATE For each sample $( \boldsymbol{x}_0, y) \in \mathcal{B}_b$, perform its corresponding backward pass over quantized weights and  activations to find the following partial derivatives 
\begin{align}
&\frac{\partial \mathcal{L}_a \Big( \boldsymbol{x}_0,y,  \{\mathsf{Q}_{\rm p}(\boldsymbol{w}), \mathsf{Q}_{\rm p}(\boldsymbol{x}) \} \Big)}{\partial \boldsymbol{\Omega}} \label{alg:eq1}
\end{align}    
at $\boldsymbol{\Omega} = (\boldsymbol{\Omega})^e_{b-1}$. 
\STATE Compute $  \frac{\partial \mathsf{H}(\boldsymbol{w}) }{\partial \boldsymbol{\Omega}}$ at $\boldsymbol{\Omega} = (\boldsymbol{\Omega})^e_{b-1}$. 

\STATE Use \eqref{eq:opt4} to compute 
\begin{align}
   & \frac{\partial \mathcal{J}_{\mathcal{B}_b} \Big(  \{\mathsf{Q}_{\rm p}(\boldsymbol{w}), \mathsf{Q}_{\rm p}(\boldsymbol{x})\}, 
 \boldsymbol{q}, \boldsymbol{s}, \boldsymbol{\alpha}, \boldsymbol{\beta} \Big)}{\partial \boldsymbol{\Omega}} \nonumber 
\end{align}
at $\boldsymbol{\Omega} = (\boldsymbol{\Omega})^e_{b-1}$. 

\STATE Update $(\boldsymbol{\Omega})^e_{b-1}$ to $(\boldsymbol{\Omega})^e_{b}$ by GD with the learning rate $\eta_w$ and  the layer-wise learning rates given in \eqref{eq:gradscale}. 
\ENDFOR
\ENDFOR
\RETURN $\boldsymbol{\Omega}^{E_{\rm max}}$.
\end{algorithmic} 
\end{algorithm}

For learning rates, we make an adjustment for each set of the trainable parameters appropriately. For updating $\boldsymbol{w}$, we use a similar $\eta_{w}$ to that of conventional DL. On the other hand, for the other four sets of parameters we should take a careful consideration. Specifically, it has been demonstrated that effective convergence occurs during training when the ratio of the average update magnitude to the average parameter magnitude remains consistent across all weight layers within a network \cite{you2017large}. In addition, we anticipate the step-size parameter to decrease with increasing precision, as finer quantization has smaller quantization step sizes. Likewise, since quantization parameters 
at layer $l$ impact all weights/activations at layer $l$, their impact on the loss function would be the aggregated impact from all weights/activations they are used to quantize. As such, their learning rates should be inversely affected by the number of  weights/activations they are used to quantize. This, together with \cite{esser2019learned}, motivates us to scale the learning rates for layer $l$, $l \in [L]$, as follows
% \small
\begin{subequations}\label{eq:gradscale}
\begin{align} \label{eq:lr_qw}
\eta_{q_l} &= \frac{1}{ \sqrt{|\boldsymbol{w}_l| 2^{b-1}}} \eta_{q}, \\ \label{eq:lr_qx}
\eta_{s_l} &= \frac{1}{ \sqrt{|\boldsymbol{x}_l| 2^{b}}} \eta_{s},\\ \label{eq:lr_alpha_w}
\eta_{\alpha_l} &= \frac{1}{ \sqrt{|\boldsymbol{w}_l|}} \eta_{\alpha},\\ \label{eq:lr_alpha_x}
\eta_{\beta_l} &= \frac{1}{ \sqrt{|\boldsymbol{x}_l|}} \eta_{\beta},
\end{align}
\end{subequations}
% \normalsize
for some $\eta_{q}, \eta_{s}, \eta_{\alpha},
\eta_{\beta} > 0$.

The proposed algorithm for optimization problem \eqref{eq:joint_opt} is summarized in Algorithm \ref{alg:01}. To simplify our notation, we use  $(\cdot)^e_{b}$ to indicate parameters of the $b$-th batch updation during the $e$-th epoch of the algorithm. We further write  $(\cdot)^e_{B}$ as $ (\cdot)^e$ whenever needed, and set $(\cdot)^e_{0}=(\cdot)^{e-1}$.

\subsection{R-CDL} \label{sec:R-CDL}

 R-CDL is almost identical to CDL except that (i) the probabilistic quantizers $\mathsf{Q}_{\rm p}(\cdot)$ are replaced with their respective \textit{soft deterministic} approximations $\mathsf{Q}_{\rm d}(\cdot)$; and (ii) forward and backward passes are operated on softly quantized weights and activations with full precision. In R-CDL, the learning process is to solve the following optimization problem 
\begin{align} \label{eq:joint_opt2}
\min_{(\boldsymbol{q}, \boldsymbol{s}, \boldsymbol{\alpha}, \boldsymbol{\beta}, \boldsymbol{w})}   ~   \mathbb{E}_{(X,Y)} \Big\{ &\mathcal{L} (X, Y, \{\mathsf{Q}_{\rm d}(\boldsymbol{w}), \mathsf{Q}_{\rm d}(\boldsymbol{x}) \} ) \nonumber \\ & \qquad + \lambda \mathsf{H}(\boldsymbol{w}) + \gamma \mathsf{H}(\boldsymbol{x}) \Big\} . 
\end{align}

Compared to CDL, R-CDL offers the advantage of computing all gradients analytically without approximation during training, albeit at the cost of increased computation for forward and backward passes using floating-point operations. This advantage enables R-CDL to achieve superior accuracy-compression trade-off compared to CDL.

\begin{remark}
In R-CDL, weights and activations are still compressible at any stage of training. They can be quantized using the newly obtained CPMFs whenever the inter-device weight and activation communication need arises in model/data parallelism; weights are still quantized at the end of training.    
\end{remark}

\begin{figure*}[!t] 
  \centering 
  \subfloat[ResNet-18 (FP accuracy:70.5).]{\includegraphics[width=0.45\linewidth]{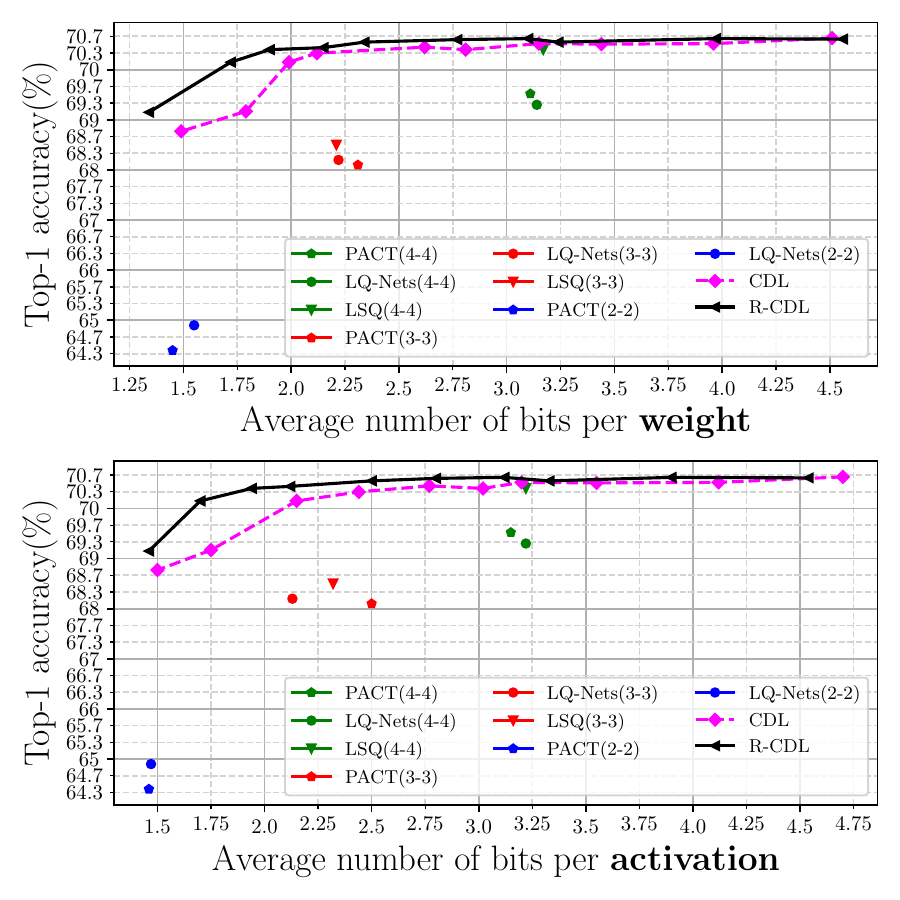}
\label{fig:ResNet_18_S}}
  \subfloat[ResNet-34 (FP accuracy:74.1).]{\includegraphics[width=0.45\linewidth]{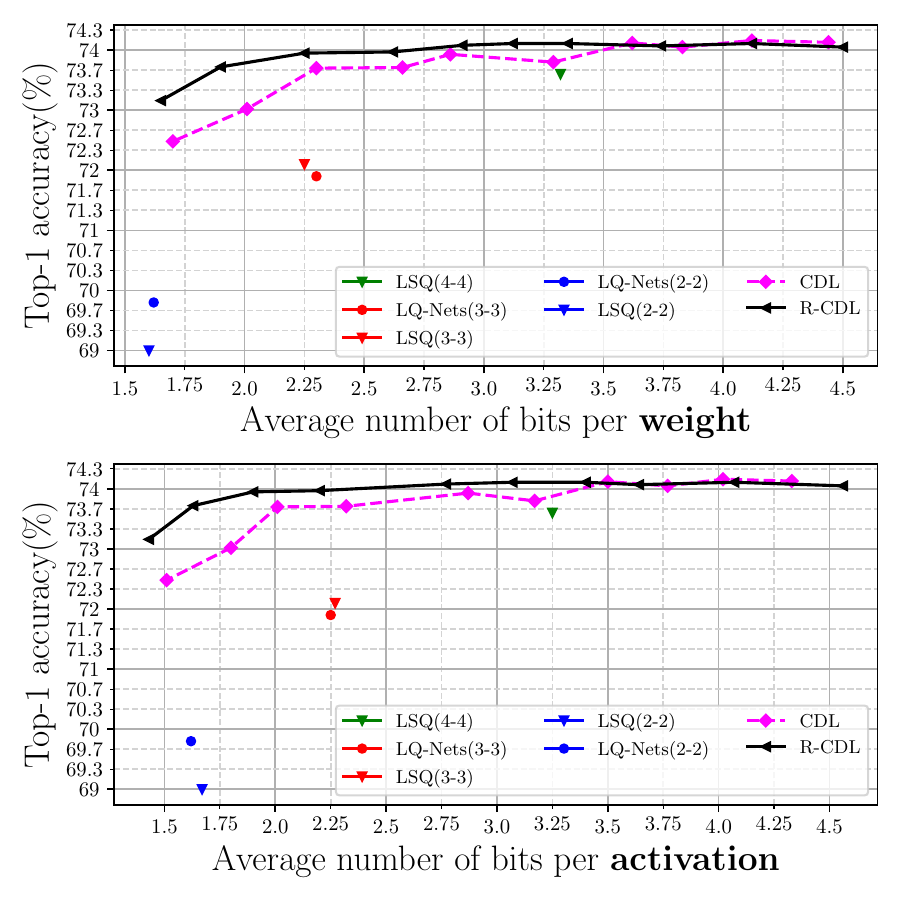}
\label{fig:ResNet_34_S}}
\vskip -0.05in
  \caption{Comparison of models trained by CDL, R-CDL, and benchmark methods in terms of the Top-1 accuracy vs the average number of bits per weight (top)/activation (bottom) on ImageNet: (a) ResNet-18, and (b) ResNet-34. All models are trained from \underline{\textbf{scratch}}.}  \vskip -0.1in \label{fig:ImageNet}
  % \vskip -0.4in 
\end{figure*}

\begin{table*}[!t]
\centering
\caption{(Top-1 accuracy, Average \# of bits per weight,  Average \# of bits per activation) given by models trained by CDL and R-CDL with $(\lambda,\gamma)=(0,0)$ and  $b = \{2,3,4\}$.} 
\vskip -0.1in
\resizebox{0.65\textwidth}{!}{\begin{tabular}{c|l|c|c|c}
\toprule[0.4mm]
\rowcolor{mygray} Model & Method & $b = 4$ & $b = 3$  & $b = 2$ \\
\midrule
\multirow{2}{*}{ResNet-18} & CDL & (70.45\%,  3.08,  3.18)  & (69.72\%,  2.25,  2.23) & (67.13\%,  1.50, 1.66) \\
&R-CDL & (70.50\%,  2.88,  2.76) & (69.95\%,  2.10,  2.12) & (67.88\%,  1.40,  1.51)\\
\midrule
\multirow{2}{*}{ResNet-34} & CDL & (73.80\%,  3.33,  3.31) & (73.01\%,  2.25,  2.34) & (70.88\%,  1.55,  1.52) \\
&R-CDL & (73.92\%,  3.04,  3.12) & (73.25\%,  2.20,  2.29) & (71.44\%,  1.45,  1.46) \\
\bottomrule[0.4mm]
\end{tabular}} \label{tab:quant}
\vskip -0.1in
\end{table*}

\begin{figure*}[!t]
  \centering 
  \subfloat[ResNet-20 (FP accuracy: 67.92\%).]{\includegraphics[width=0.45\linewidth]{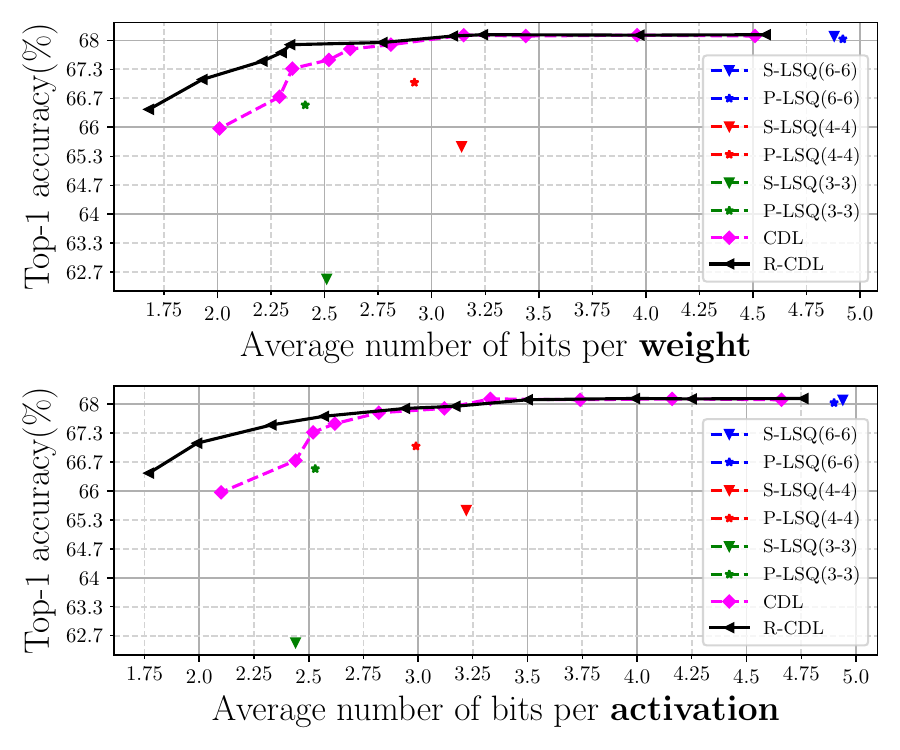}
\label{fig:ResNet-20}}
  % \hfill
  \subfloat[ResNet-44 (FP accuracy: 71.85\%).]{\includegraphics[width=0.45\linewidth]{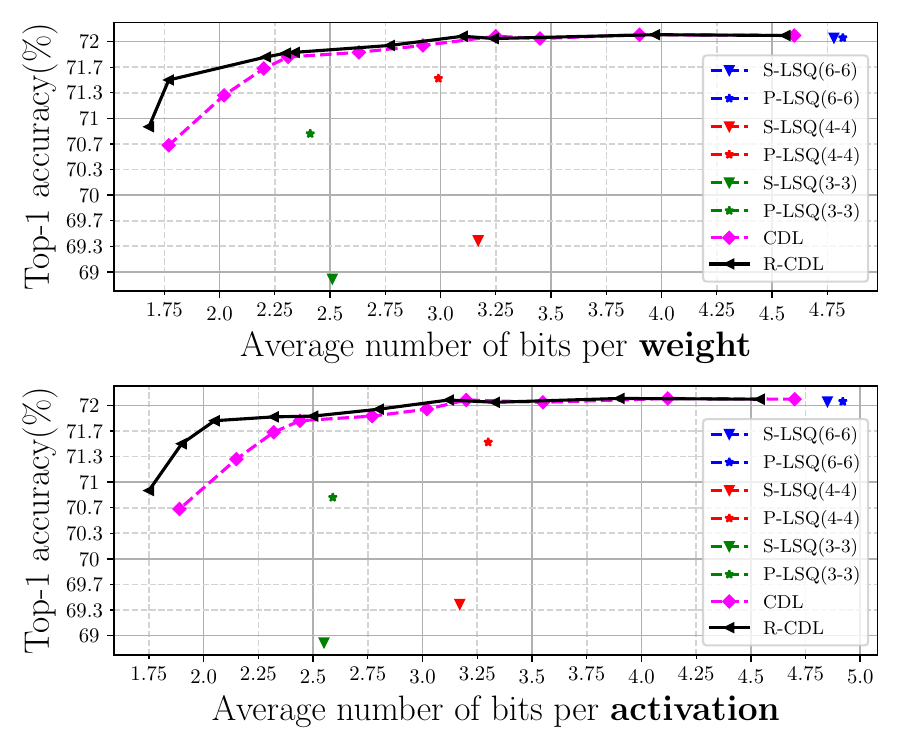}
\label{fig:ResNet-44}}
  % \hfill
  \\ \vskip -0.05in
  \subfloat[ResNet-56 (FP accuracy: 72.05\%).]{\includegraphics[width=0.45\linewidth]{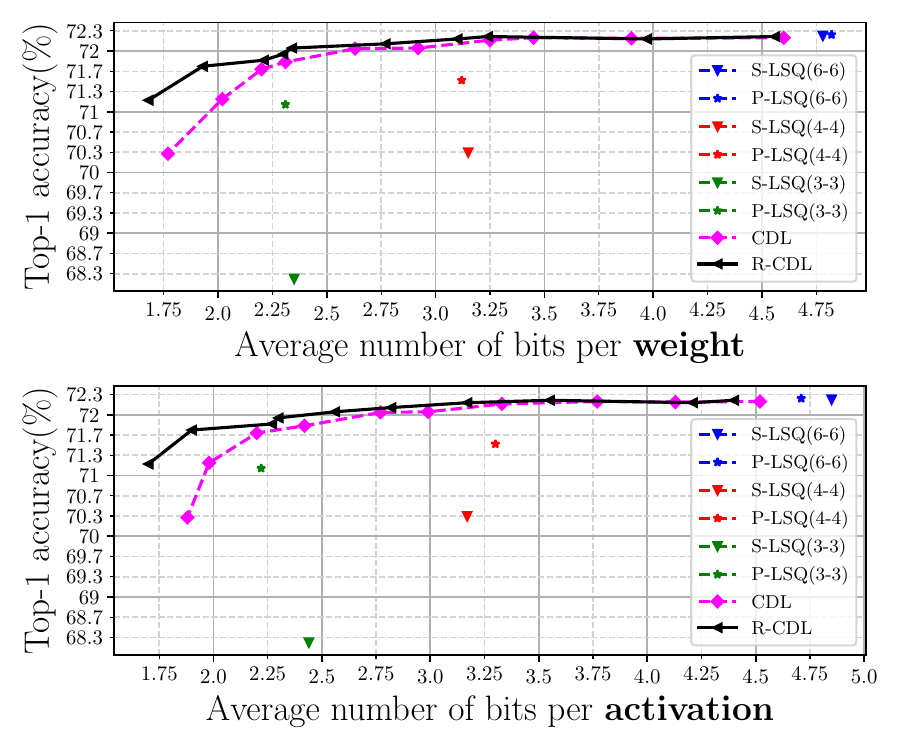}
\label{fig:ResNet-56}}
  \subfloat[ResNet-110 (FP accuracy: 73.04\%).]{\includegraphics[width=0.45\linewidth]{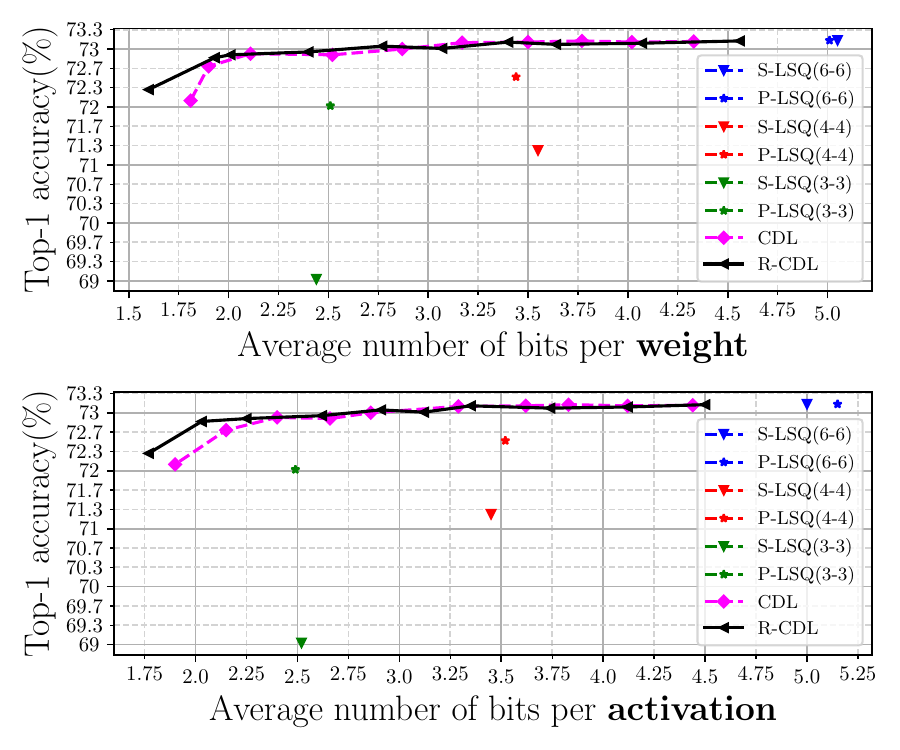}
\label{fig:ResNet-110}} \\ \vskip -0.05in
  \subfloat[[VGG13 (FP: 73.70\%).]{\includegraphics[width=0.45\linewidth]{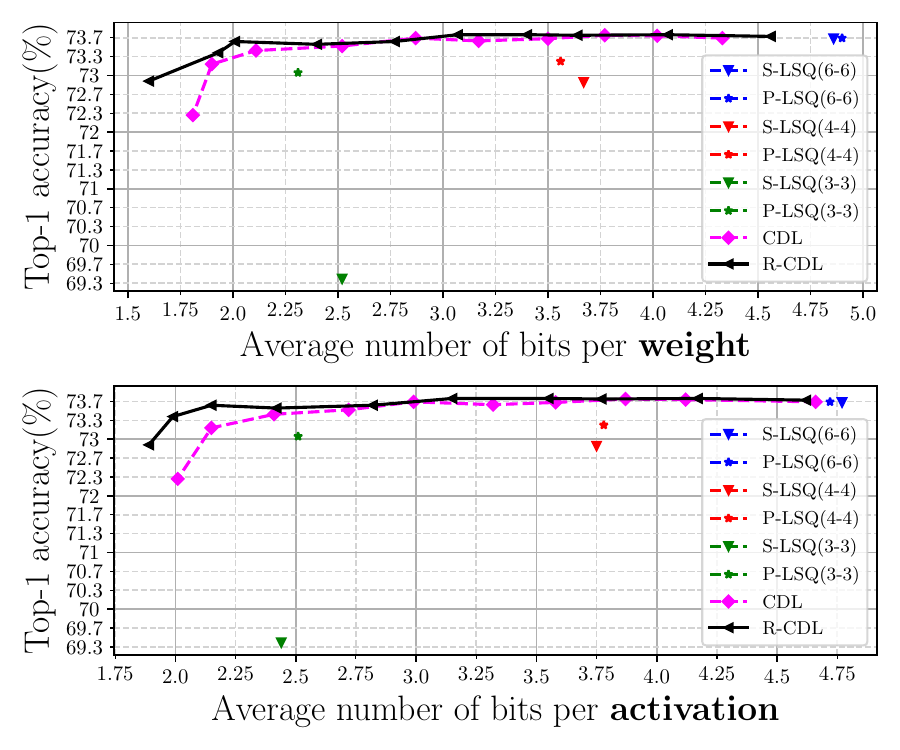}
\label{fig:VGG13}}
  \subfloat[WRN-28-10 (FP: 80.88\%).]{\includegraphics[width=0.45\linewidth]{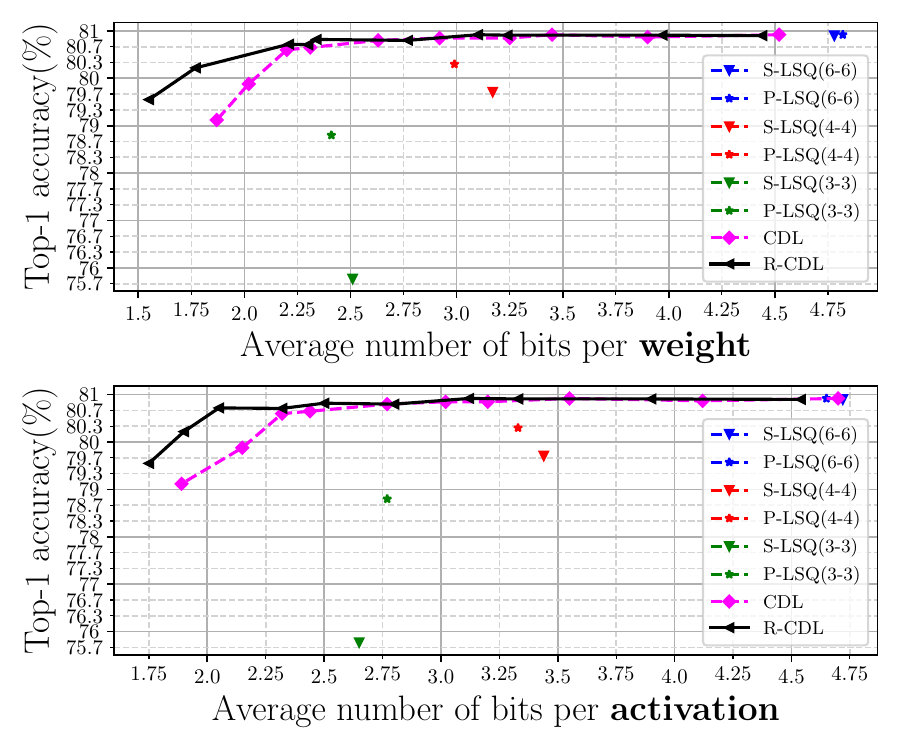}
\label{fig:wrn}}
  \caption {Comparison of models trained by CDL, R-CDL, and benchmark methods in terms of the test accuracy vs the average number of bits per weight (top)/activation (bottom) on CIFAR-100 dataset: (a) ResNet-20, (b) ResNet-44, (c) ResNet-56, (d) ResNet-110, (e) VGG-13, and (e) Wide-ResNet-28-10.}
  \label{fig:cifar100}
  \vspace{-3mm}
\end{figure*}

\section{Experiments} \label{sec:exp}
To demonstrate the effectiveness of CDL and R-CDL and compare them with existing SOTA methods in the literature, we have conducted a comprehensive series of experiments. This section presents these experiment results. We begin with 
 defining performance metrics to be used to compare CDL and R-CDL with the benchmark methods.

\subsection{Performance Metrics} \label{sec:metric}
In all cases, we report the test accuracy of each trained DNN vs the average number of bits required to represent the weights/activations of the DNN, where the latter is defined as
\begin{subequations} \label{eq:Nofbits}
\begin{align} 
\text{average \# of bits per weight} &= \frac{\sum_{l=1}^L |\boldsymbol{w}_l| b_{\boldsymbol{w}_l}}{\sum_{l=1}^L |\boldsymbol{w}_l|}, \\
\text{average \# of bits per activation} &= \frac{\sum_{l=1}^{L-1}  |\boldsymbol{x}_l| b_{\boldsymbol{x}_l}}{\sum_{l=1}^{L-1} |\boldsymbol{x}_l|} ,
\end{align}
\end{subequations}
where $b_{\boldsymbol{w}_l}$ and $b_{\boldsymbol{x}_l}$ denote the average number of bits per weight and activation, respectively, required to represent the $l$-th layer of the DNN using Huffman coding. Particularly, to calculate $b_{\boldsymbol{x}_l}$, we first perform an inference over the samples in a mini-batch of the training dataset to obtain the respective $\boldsymbol{x}_l$ for these samples, and then encode these activations using Huffman coding.

Furthermore, for a fair comparison with the benchmark methods, we always apply Huffman coding to encode the weights and activations of the fully-trained models obtained using the benchmark methods.

\subsection{Implementation Details of CDL and R-CDL} \label{sec:impdetails}
In this subsection, we delve into the implementation details of CDL and R-CDL which is used throughout this section. 

\subsubsection{Quantization operations in a mini-batch} As the softmax operation is time-consuming, deploying quantizers $\mathsf{Q}_{\rm p}(\cdot)$ and $\mathsf{Q}_{\rm d}(\cdot)$ incurs some computation overheads. This will not be an issue when deploying these quantizers for the weights as the weights are quantized only once per mini-batch (refer to Algorithm \ref{alg:01}). On the other hand, when quantizing the activations, these quantizers should be applied over all the samples in a mini-batch, which incurs time complexity. To address this issue, for quantizers $\mathsf{Q}_{\rm p}(\cdot)$ and $\mathsf{Q}_{\rm d}(\cdot)$ used for activations, each of their corresponding CPMFs $\big[P_{\beta_l}( \cdot | \boldsymbol{x}_l[j])\big]$ is modified to be concentrated proportionally on the top five values with the rest of conditional probabilities to be $0$.

\subsubsection{First and last layer quantization}
We quantize all weights to a low bit-width, with the exception of the weights in the first and last layers, which are kept at 8-bits, similar to the approach used in benchmark methods \cite{lsq, lsq+, lq-net}.

\subsubsection{Common hyper-parameters of Algorithm 1}
For all performance curves reported in the paper,  we always set $b=6$ for both weights and activations in Algorithm \ref{alg:01}, and test different $\lambda$ and $\gamma$ values. For all the five sets of parameters $\{\boldsymbol{q}, \boldsymbol{s}, \boldsymbol{\alpha}, \boldsymbol{\beta}, \boldsymbol{w}\}$, we use a similar optimizer, and the same values for the learning rates $\{\eta_{w}, \eta_{q}, \eta_{s}, \eta_{\alpha},
\eta_{\beta}\}$. Note that the learning rates for quantization parameters are further adaptively scaled layer-wise during training according to \eqref{eq:gradscale}.

Inspired from \cite{lsq}, we set $\boldsymbol{q}^0 = \frac{2 \overline{|\boldsymbol{w}|}}{\sqrt{2^{b-1}}}$ and $\boldsymbol{s}^0 = \frac{2 \overline{|\boldsymbol{x}|}}{\sqrt{2^{b-1}}}$, where $\overline{|\boldsymbol{w}|}$ is the average of absolute values of initial weights, and $\overline{|\boldsymbol{x}|}$ is the average of absolute values for the first batch of activations. Also, we set $\boldsymbol{\alpha}^0 = [500]_L$ and $ \boldsymbol{\beta}^0 = [500]_{L-1}$.

\subsection{Experiments on ImageNet} \label{sec:imagenet}

ImageNet is a large-scale dataset containing around 1.2 million training samples and 50,000 validation images.

$\bullet$ \textbf{Models}: 
We have conducted experiments on two models from ResNet family, namely ResNet-18 and ResNet-34 \cite{he2016deep}.

$\bullet$ \textbf{Benchmarks}: We evaluate the performance of CDL (and R-CDL) against several QAT methods, including LSQ \cite{lsq}, PACT \cite{pact2018}, LQ-Nets \cite{lq-net}. In this subsection, all the benchmark methods, including CDL, are trained from \textbf{scratch}.  Additionally, in Appendix \ref{app:imagenet}, we explore the effect of initializing training from pre-trained models on validation accuracy, where we also add two additional  benchmarks, namely APoT \cite{apot} and DMBQ \cite{zhao2021distribution}. These two benchmarks are not included in the current subsection due to their poorer performance when trained from scratch.

$\bullet$ \textbf{Training settings}: 
We have deployed an SGD optimizer with a momentum
of 0.9, a weight decay of 0.0001, and a batch size of 256. We have trained the models for 90 epochs, and adopted an initial learning rate of 0.1, which is further divided  by 10 at the 30-th and 60-th epochs. 

We have run Algorithm \ref{alg:01} for different pairs of $\lambda$ and $\gamma$ values $(\lambda,\gamma) =\{(0,0), (0.01,0.01), \dots, (0.09,0.09)\}$ .

$\bullet$ \textbf{Results}: 
Fig. \ref{fig:ImageNet} shows the performance of the Top-1 accuracy vs the average number of bits per weight (top)/activation (bottom) for models trained by CDL, R-CDL, and benchmark methods. Specifically, these curves are depicted for ResNet-18 and ResNet-34 in Figs. \ref{fig:ResNet_18_S} and \ref{fig:ResNet_34_S}, respectively. 

To demonstrate the performance gain delivered only by  the proposed quantizers $\mathsf{Q}_{\rm p}(\cdot)$ and $\mathsf{Q}_{\rm d}(\cdot)$, we have also run Algorithm \ref{alg:01} with $(\lambda,\gamma)=(0,0)$, and  $b=\{2,3,4\}$. The corresponding results for both ResNet-18 and ResNet-34 on ImageNet are listed in Table \ref{tab:quant}.

By examining the curves in Fig. \ref{fig:ImageNet} and the numbers in Table \ref{tab:quant}, it is fair to make the following observations:
\begin{itemize}
    \item \textit{Observation 1}:  The accuracy of the models trained by R-CDL is always above those trained by CDL. This is because unlike CDL, the gradient calculations in R-CDL are performed with no approximation.

    \item \textit{Observation 2}: The models trained by CDL and R-CDL  can even have slightly higher test accuracy compared to the respective pre-trained FP models while achieving significant compression. For example, ResNet-18 trained by R-CDL has 70.55\% accuracy (slightly better than 70.5\% FP accuracy) with 2.34 bits per weight, achieving more than 13 fold compression; likewise, ResNet-18 trained by CDL has 70.52\% accuracy (slightly better than 70.5\% FP accuracy) with 3.34 bits per weight, achieving more than 9 fold compression. Similar results are valid for ResNet-34 as well.
    
    \item \textit{Observation 3}: In the case of low bit rates,  the accuracy of the models trained via CDL and R-CDL is significantly better than that of those trained by benchmark methods.

    \item \textit{Observation 4}: The proposed quantizers $\mathsf{Q}_{\rm p}(\cdot)$ and $\mathsf{Q}_{\rm d}(\cdot)$ on their own also outperforms benchmark quantization methods in terms of model test accuracy. 
\end{itemize}

% \begin{table}[h]
% \centering
% \caption{Top-1 accuracy (\%) of models trained by CDL and R-CDL when they are initialized with different number of bits, while $(\lambda,\gamma)=(0,0)$.} 
% \vskip -0.1in
% \resizebox{1\columnwidth}{!}{\begin{tabular}{c|l|c|c|c|c|c}
% \toprule[0.4mm]
% \rowcolor{mygray} Model & Method & $b = 6$  & $b = 5$& $b = 4$ & $b = 3$  & $b = 2$ \\
% \midrule
% \multirow{2}{*}{ResNet-18} & CDL & 70.63 & 70.62 & 70.45 & 69.72 & 67.13 \\
% &R-CDL & 70.61 & 70.63 & 70.50 & 69.95 & 67.88\\
% \midrule
% \multirow{2}{*}{ResNet-34} & CDL & 74.13 & 74.02 & 73.80 & 73.01 & 70.88 \\
% &R-CDL & 74.05 & 74.12 & 73.92 & 73.25 & 71.44\\
% \bottomrule[0.4mm]
% \end{tabular}} \label{tab:quant}
% \end{table}

\begin{figure*}[!t]
\centering  \includegraphics[width=1\textwidth]{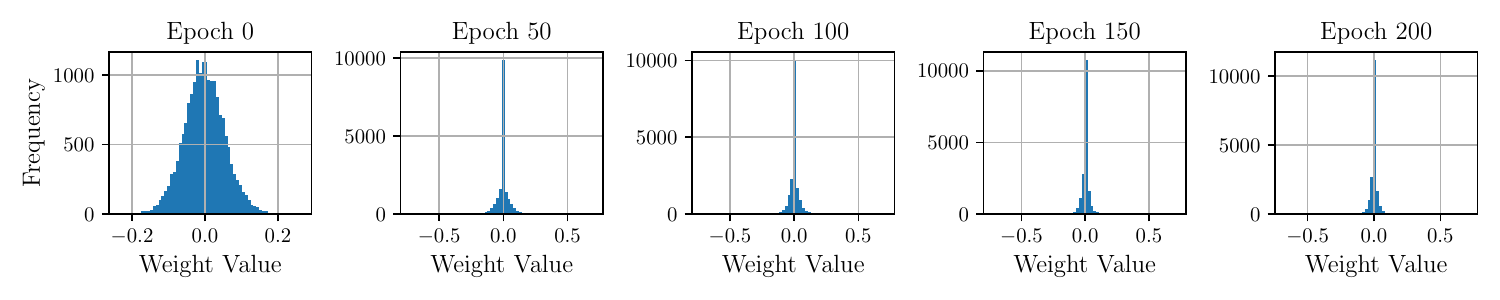}
\vskip -0.2in
  \caption{The evolution of weight distribution of a convolutional layer in ResNet-110 during training via CDL on CIFAR-100.
} \vskip -0.2in
\label{fig:distweight}
\end{figure*}

\subsection{Experiments on CIFAR-100} \label{sec:cifar}
CIFAR-100 dataset contains 50K training and 10K test images of size $32\times32$, which are labeled for 100 classes.

$\bullet$ \textbf{Models}: 
To demonstrate the effectiveness of CDL and R-CDL, we have conducted experiments using three different model architectural families. Specifically, we have selected (i) four models from the ResNet family \cite{he2016deep}, namely ResNet-$\{20, 44, 56, 110\}$; (ii) VGG-13 from the VGG family \cite{simonyan2014very}; and (iii) Wide-ResNet-28-10 from the Wide-ResNet family \cite{zagoruyko2016wide}. 

$\bullet$ \textbf{Benchmarks}: We have trained the selected DNN models from scratch via CDL and R-CDL. They are evaluated against  those trained via LSQ either on top of pre-trained models (denoted as P-LSQ) or from scratch (denoted as S-LSQ).

$\bullet$ \textbf{Training settings}: 
 We have deployed an SGD optimizer with a momentum
of 0.9, a weight decay of 0.0005, and a batch size of 64. We have trained the models for 200 epochs, and adopted an initial learning rate of 0.1, which is further divided by
10 at the 60-th, 120-th and 160-th epochs. To have a fair comparison,  we have reproduced the results of LSQ on our local machine.

We have run Algorithm \ref{alg:01} for different pairs of $\lambda$ and $\gamma$ values $(\lambda,\gamma) =\{(0,0), (0.01,0.01), \dots, (0.09,0.09)\}$ .

$\bullet$ \textbf{Results}: The results are presented in Fig. \ref{fig:cifar100}. Specifically, the results for ResNet-20, ResNet-44, ResNet-56, ResNet-110, VGG-13, and Wide-ResNet-28-10  are illustrated in Figs. \ref{fig:ResNet-20}, \ref{fig:ResNet-44}, \ref{fig:ResNet-56}, \ref{fig:ResNet-110}, \ref{fig:VGG13}, and \ref{fig:wrn}, respectively. Again, the observations made in Subsection~\ref{sec:imagenet} for ImageNet remain valid as well for CIFAR-100. Additionally, we note that, generally, the accuracy gap between CDL and R-CDL diminishes as the model size increases.

\subsection{Evolution of Weight Distribution} \label{sec:wevolve}
Since the entropy term $\mathsf{H}(\boldsymbol{w})$ (and $\mathsf{H}(\boldsymbol{x})$) in the optimization problem \eqref{eq:joint_opt} imposes certain structures on the weights (and activations), it is interesting to see how CDL changes the distribution of the weights in favour of higher compression ratio. Fig. \ref{fig:distweight} illustrates the evolution of the distribution of flattened weights in a convolutional layer of ResNet-110 during the training process within CDL on CIFAR-100.  Specifically, $b=6$, and the weight distribution is depicted for epochs 0, 50, 100, 150 and 200 using $2^6$ bins. As observed, over time, the distribution of weights becomes more concentrated around the center, leading to reduced entropy. This reduction in entropy in turn leads to a higher compression ratio when employing Huffman coding.

\subsection{Average Number of Bits vs Epoch for Facilitating Data/Model Parallelism} \label{sec:avgbits}

\begin{figure*}[!t]
  \centering 
  \subfloat[CDL Weights.]{\includegraphics[width=0.45\linewidth]{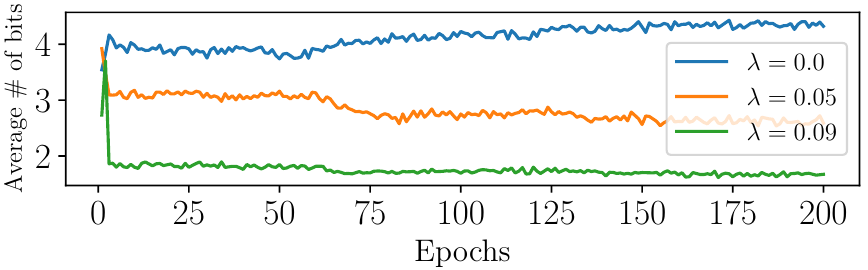}
\label{fig:bits_w}}
  % \hfill
  \subfloat[CDL Activations.]{\includegraphics[width=0.45\linewidth]{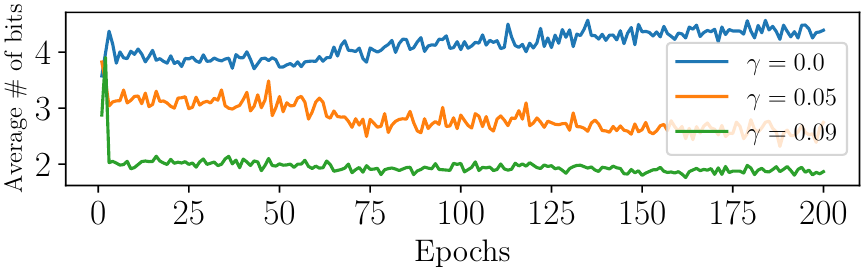}
\label{fig:bits_a}} \\ \vskip -0.1in
  \subfloat[R-CDL Weights.]{\includegraphics[width=0.45\linewidth]{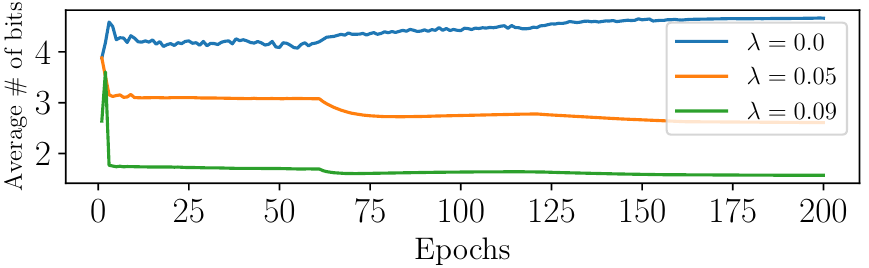}
\label{fig:R-CDL-w}}
  % \hfill
  \subfloat[R-CDL Activations.]{\includegraphics[width=0.45\linewidth]{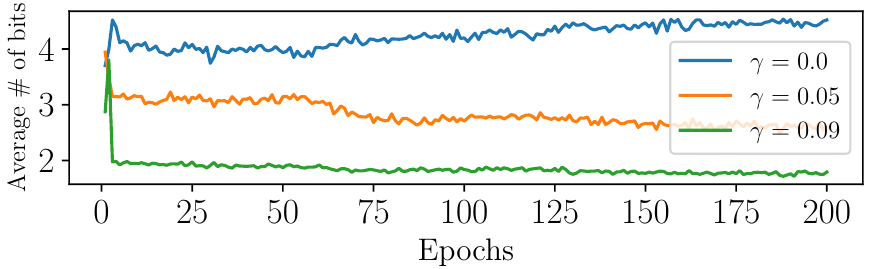}
\label{fig:R-CDL-a}}
\vskip -0.05in
  \caption{The average number of bits required to encode (a) CDL weights, (b) CDL activations, (c) R-CDL weights, and (d) R-CDL activations vs epochs for ResNet-110 trained over CIFAR-100.}
  \label{fig:CDL_bits}
  \vskip -0.2in 
\end{figure*}

To show that weights and activations are indeed compressible at any stage of training within CDL and R-CDL, 
we depict the average number of bits required to encode the weights and activations of ResNet-110 trained via CDL or R-CDL over the course of training on CIFAR-100 in Fig.~\ref{fig:CDL_bits}. We calculate the average number of bits as follows. In the case of CDL, since the weights and activations are quantized at each iteration step during training, we proceed to encode them after the completion of each epoch using Huffman coding. For R-CDL, after each epoch is finished,  we use $\mathsf{Q}_{\rm p}(\cdot)$  with the newly obtained quantization parameters to quantize the weights and activations, and then encode the quantized values using Huffman coding. 

The results for CDL weights and activations are shown in Figs. \ref{fig:bits_w} and \ref{fig:bits_a}, respectively, while those for R-CDL weights and activations are displayed in Figs. \ref{fig:R-CDL-w} and \ref{fig:R-CDL-a}, respectively. As seen, for $\lambda, \gamma >0$, the average number of bits  drops quickly after a few epochs. This in turn would reduce the inter-device weight and activation communication costs quickly in  model/data parallelism.

\subsection{Convergence Curves for CDL and R-CDL} \label{sec:conv}

\begin{figure*}[!t]
  \centering 
  \vskip -0.1in
  \subfloat[CDL.]{\includegraphics[width=0.45\linewidth]{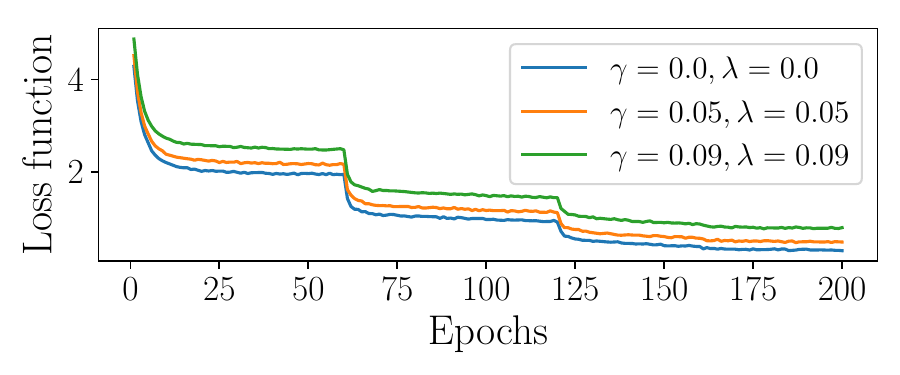}
\label{fig:conv_CDL}}
  % \hfill
  \subfloat[R-CDL.]{\includegraphics[width=0.45\linewidth]{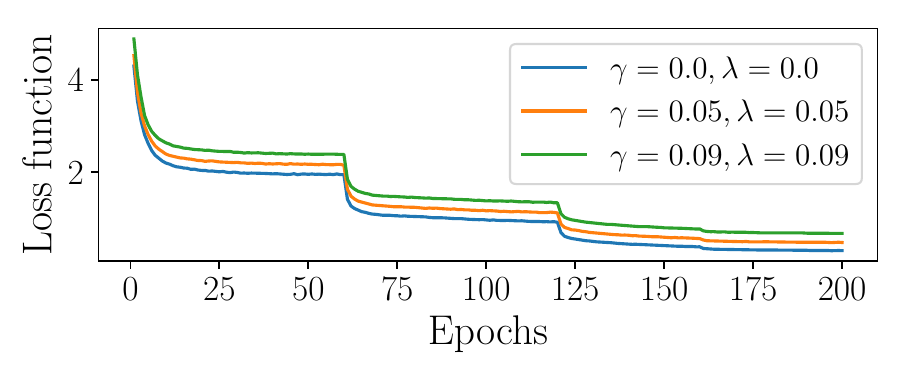}
\label{fig:conv_RCDL}}
\vskip -0.05in
  \caption{The convergence of objective function in (a) CDL, and (b) R-CDL over the course of training ResNet-110 on CIFAR-100.}
  \label{fig:conv}
  \vskip -0.1in 
\end{figure*}

To illustrate the convergence of  Algorithm \ref{alg:01} in both cases of CDL and R-CDL, we plot the evolution of the objective function of CDL given in \eqref{eq:joint_opt} and of R-CDL given in \eqref{eq:joint_opt2} over the course of training ResNet-110 on CIFAR-100 in Fig.~\ref{fig:conv}. Particularly, the curves for CDL and R-CDL are depicted in Figs. \ref{fig:conv_CDL} and \ref{fig:conv_RCDL}, respectively. As observed, both curves flatten out near the end of the training, indicating that the algorithm has converged in each case.

\section{Conclusion}\label{Sec:conclusion}
In this paper, we have introduced a novel probabilistic quantization method and its soft deterministic variant. Empowered by these quantization methods, we have then developed ``coded deep learning''' (CDL), a new framework that harnesses information-theoretic coding techniques in the training and inference of deep neural network (DNN) model to achieve substantial compression of model weights and activations, while maintaining or even improving model accuracy, reducing computational complexity at both training and inference stages, and enabling efficient implementation of model/data parallelism. Key elements of CDL include: (1) incorporating trainable probabilistic quantizers to quantize both model weights and activations; (2) enforcing entropy constraints on quantized weights and activations during training, and (3) using the soft deterministic counterparts of the trainable probabilistic quantizers as a proxy to facilitate gradient computation in backpropagation. A variant of CDL called  ``Relaxed CDL'' (R-CDL) has also been proposed to further improve the trade-off between validation accuracy and compression at the disadvantage of full precision operation in both forward and backward passes during training. Extensive experimental results have demonstrated that both CDL and R-CDL outperform state-of-the-art methods in the field of DNN compression, with significant gains in the low bit rate case.

\appendices
\section{Proof of Proposition \ref{prop:distortion}}
\label{app:dist}
Note that in view of \eqref{eq:Qprob} and \eqref{eq:softQ}, 
   \begin{equation} \label{eq:a1}
       \mathsf{Q}_{\rm d}(\theta) = \mathbb{E} \big[  \mathsf{Q}_{\rm p}(\theta)| \theta \big ]. 
   \end{equation}
Then it follows that
\begin{align} \label{eq:appEdistort}
&\mathbb{E} \big [  (\theta - \mathsf{Q}_{\rm p}(\theta) )^2 ~ | ~ \theta  \big ] \\
&=\mathbb{E} \big [  \big( \theta -  \mathsf{Q}_{\rm d}(\theta)  + \mathsf{Q}_{\rm d}(\theta)  - \mathsf{Q}_{\rm p}(\theta) \big)^2 | \theta \big ] \\ \nonumber
&=\mathbb{E} \big [\big( \theta - \mathsf{Q}_{\rm d}(\theta)   \big)^2 | \theta \Big]  + \mathbb{E} \big [  \big( \mathsf{Q}_{\rm p}(\theta) - \mathsf{Q}_{\rm d}(\theta)  \big)^2 | \theta  \big ] \\
&\qquad + 2 \mathbb{E} \big [  (\theta - \mathsf{Q}_{\rm d}(\theta)  ) (\mathsf{Q}_{\rm d}(\theta)   - \mathsf{Q}_{\rm p}(\theta)) | \theta  \big ] \\
&=  \big( \theta  - \mathsf{Q}_{\rm d}(\theta)  \big)^2 + \text{Var}\big\{\mathsf{Q}_{\rm p}(\theta)~|~\theta \big\} \nonumber \\ 
&+ 2 (\theta - \mathsf{Q}_{\rm d}(\theta)  )  \mathbb{E} \big [  ( \mathsf{Q}_{\rm d}(\theta)  - \mathsf{Q}_{\rm p}(\theta)) | \theta  \big ] \nonumber \\
& = \big( \theta  - \mathsf{Q}_{\rm d}(\theta)  \big)^2 + \text{Var}\big\{\mathsf{Q}_{\rm p}(\theta)~|~\theta \big\}, \label{eq:distproof}
\end{align} 
where the last equality  \eqref{eq:distproof} is due to \eqref{eq:a1}. This completes the proof of Proposition \ref{prop:distortion}.

\section{Proof of Lemma \ref{lemma:partial}} \label{app:partial}
To find $\frac{\partial \mathsf{Q}_{\rm d}(\theta)}{\partial \theta}$, we have
\small
\begin{align} \label{eq:partialw1}
\frac{\partial \mathsf{Q}_{\rm d}(\theta)}{\partial \theta}  = \sum_{i \in \mathcal{A}} iq \frac{ \partial P_{\alpha}(iq|\theta)}{\partial \theta}  
\end{align}
\normalsize
To continue, it follows that
\small
\begin{align} 
& \frac{ \partial P_{\alpha}(iq|\theta)}{\partial \theta} = \frac{-2 \alpha e^{-\alpha (\theta - iq)^2} (\theta - iq)}{ \sum_{j \in \mathcal{A}} e^{-\alpha (\theta - jq)^2}} \nonumber \\
& \qquad \qquad \qquad + \frac{e^{-\alpha (\theta - iq)^2} \sum_{j \in \mathcal{A}} 2\alpha e^{-\alpha (\theta - jq)^2} (\theta - jq)}{\big( \sum_{j \in \mathcal{A}} e^{-\alpha (\theta - jq)^2} \big)^2} \nonumber \\ \label{eq:partialw2}
& = -2\alpha P_{\alpha}(iq|\theta) (\theta - iq) + 2\alpha P_{\alpha}(iq|\theta)  \sum_{j \in \mathcal{A}} (\theta - jq) P_{\alpha}(jq|\theta).
\end{align}
\normalsize
Then, by plugging \eqref{eq:partialw2} into \eqref{eq:partialw1}, we obtain
\small
\begin{align}
&\frac{\partial \mathsf{Q}_{\rm d}(\theta)}{\partial \theta} = -2\alpha q \sum_{i \in \mathcal{A}} i(\theta - iq)  P_{\alpha}(iq|\theta) \nonumber \\
& \quad + 2\alpha q \Big( \sum_{i \in \mathcal{A}} i P_{\alpha}(iq|\theta)  \Big) \Big( \sum_{j \in \mathcal{A}} (\theta - jq) P_{\alpha}(jq|\theta) \Big) \nonumber \\
&= -2\alpha q \theta \sum_{i \in \mathcal{A}} i P_{\alpha}(iq|\theta) + 2\alpha \sum_{i \in \mathcal{A}} (iq)^2 P_{\alpha}(iq|\theta) \nonumber \\
&\quad + 2\alpha q \Big( \sum_{i \in \mathcal{A}} i P_{\alpha}(iq|\theta)  \Big) \Big( \theta - \sum_{j \in \mathcal{A}} jq P_{\alpha}(jq|\theta)\Big)  \nonumber \\
&= 2\alpha \sum_{i \in \mathcal{A}} (iq)^2 P_{\alpha}(iq|\theta) \nonumber \\
& \quad -2\alpha \Big( \sum_{i \in \mathcal{A}} iq P_{\alpha}(iq|\theta) \Big) \Big( \sum_{j \in \mathcal{A}} jq P_{\alpha}(jq|\theta) \Big) \nonumber \\
& = 2\alpha \mathbb{E} \Big\{ \Big( \mathsf{Q}_{\rm p}(\theta) \Big)^2 \Big\} -2\alpha \Big( \mathbb{E} \big\{ \mathsf{Q}_{\rm p}(\theta) \big\} \Big)   \Big( \mathbb{E} \big\{ \mathsf{Q}_{\rm p}(\theta) \big\} \Big) \nonumber \\ 
& = 2\alpha \text{Var}  \big\{ \mathsf{Q}_{\rm p}(\theta) \big\}.
\end{align}
\normalsize
On the other hand, to find $\frac{\partial \mathsf{Q}_{\rm d}(\theta)}{\partial q}$, we have
\begin{align} \label{eq:partialq1}
\frac{\partial \mathsf{Q}_{\rm d}(\theta)}{\partial q} =   \sum_{i \in \mathcal{A}} i P_{\alpha}(iq|\theta)  + \sum_{i \in \mathcal{A}} iq \frac{ \partial P_{\alpha}(iq|\theta)}{\partial q}.
\end{align}
In addition, 
\small
\begin{align}
&\frac{ \partial P_{\alpha}(iq|\theta)}{\partial q} = \frac{2 \alpha e^{-\alpha (\theta - iq)^2} i(\theta - iq)}{ \sum_{j \in \mathcal{A}} e^{-\alpha (\theta - jq)^2}} \nonumber \\ 
& \qquad \qquad \qquad -  \frac{e^{-\alpha (\theta - iq)^2} \sum_{j \in \mathcal{A}} 2\alpha e^{-\alpha (\theta - jq)^2} j(\theta - jq)}{\big( \sum_{j \in \mathcal{A}} e^{-\alpha (\theta - jq)^2} \big)^2} \nonumber \\ \label{eq:partialq2}
& = 2\alpha i P_{\alpha}(iq|\theta) (\theta - iq) - 2\alpha P_{\alpha}(iq|\theta)  \sum_{j \in \mathcal{A}} j (\theta - jq) P_{\alpha}(jq|\theta).
\end{align}
\normalsize
Plugging \eqref{eq:partialq2} in \eqref{eq:partialq1} yields
\small
\begin{align}
&\frac{\partial \mathsf{Q}_{\rm d}(\theta)}{\partial q} = \sum_{i \in \mathcal{A}} i P_{\alpha}(iq|\theta) +2\alpha q \sum_{i \in \mathcal{A}} i^2 (\theta - iq)  P_{\alpha}(iq|\theta) \nonumber \\
& \quad - 2\alpha q \Big( \sum_{i \in \mathcal{A}} i P_{\alpha}(iq|\theta)  \Big) \Big( \sum_{j \in \mathcal{A}} j(\theta - jq) P_{\alpha}(jq|\theta) \Big) \nonumber \\
&= \sum_{i \in \mathcal{A}} i P_{\alpha}(iq|\theta) +2\alpha q \theta \sum_{i \in \mathcal{A}} i^2  P_{\alpha}(iq|\theta) - 2\alpha q^2 \sum_{i \in \mathcal{A}} i^3 P_{\alpha}(iq|\theta) \nonumber \\
& - 2\alpha q \theta \Big( \sum_{i \in \mathcal{A}} i P_{\alpha}(iq|\theta)  \Big) \Big( \sum_{j \in \mathcal{A}} j P_{\alpha}(jq|\theta) \Big) \nonumber \\
&\quad + 2\alpha q^2 \Big( \sum_{i \in \mathcal{A}} i P_{\alpha}(iq|\theta)  \Big) \Big( \sum_{j \in \mathcal{A}} j^2 P_{\alpha}(jq|\theta)  \Big)  \nonumber \\
&= \sum_{i \in \mathcal{A}} i P_{\alpha}(iq|\theta) + 2\alpha q \theta \Big( \sum_{i \in \mathcal{A}} i^2 P_{\alpha}(iq|\theta) -\big( \sum_{i \in \mathcal{A}} i P_{\alpha}(iq|\theta)\big)^2 \Big) \nonumber \\
&\quad - 2\alpha q^2  \Big(  \sum_{i \in \mathcal{A}} i^3 P_{\alpha}(iq|\theta) -\big( \sum_{i \in \mathcal{A}} i P_{\alpha}(iq|\theta)\big) \big( \sum_{i \in \mathcal{A}} i^2 P_{\alpha}(iq|\theta) \big) \Big) \nonumber \\
& = \frac{1}{q} \left ( \mathbb{E} \big\{ \mathsf{Q}_{\rm p}(\theta) \big\} + 2\alpha  \theta \text{Var}  \big\{ \mathsf{Q}_{\rm p}(\theta) \big\} -2\alpha  \text{Skew}_{\rm u} \big\{ \mathsf{Q}_{\rm p}(\theta) \big\} \right ),
\end{align}
\normalsize
which concludes the proof. 

\begin{figure*}[!t]
  \centering 
  \subfloat[ResNet-18 (FP accuracy:70.5).]{\includegraphics[width=0.45\linewidth]{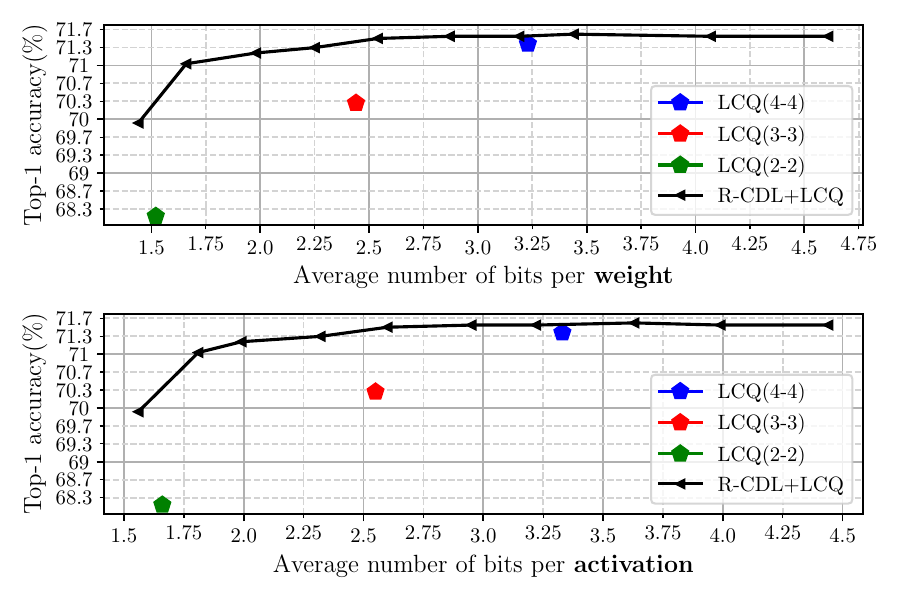}
\label{fig:non18}}
  \subfloat[ResNet-34 (FP accuracy:74.1).]{\includegraphics[width=0.45\linewidth]{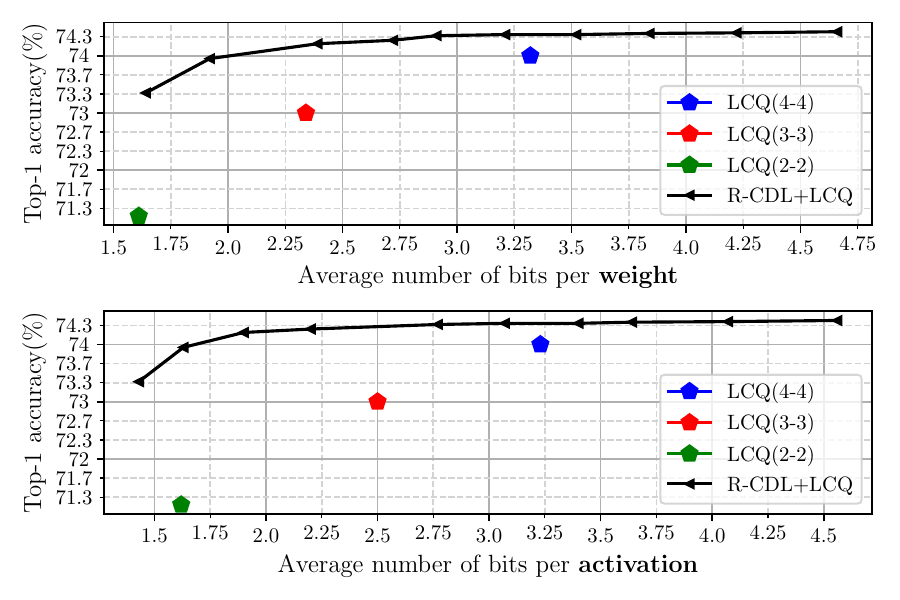}
\label{fig:non34}}
\vskip -0.05in
  \caption{Top-1 accuracy vs the average number of bits per weight (top)/activation (bottom) for LCQ, and R-CDL+LCQ on ImageNet: (a) ResNet-18, and (b) ResNet-34. All models are trained from \underline{\textbf{scratch}}.} \label{fig:non-uniform}
  \vskip -0.2in 
\end{figure*}

\begin{figure*}[!t]
  \centering 
  \subfloat[ResNet-18 (FP accuracy:70.5).]{\includegraphics[width=0.45\linewidth]{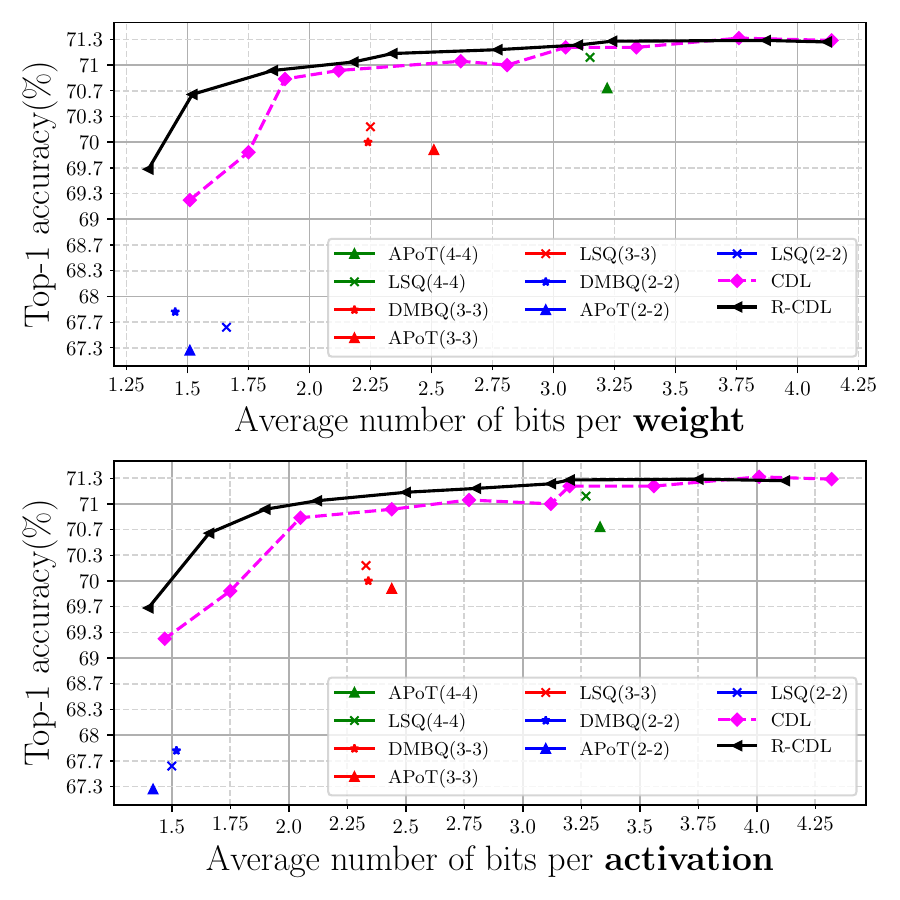}
\label{fig:ResNet-18_P}}
  \subfloat[ResNet-34 (FP accuracy:74.1).]{\includegraphics[width=0.45\linewidth]{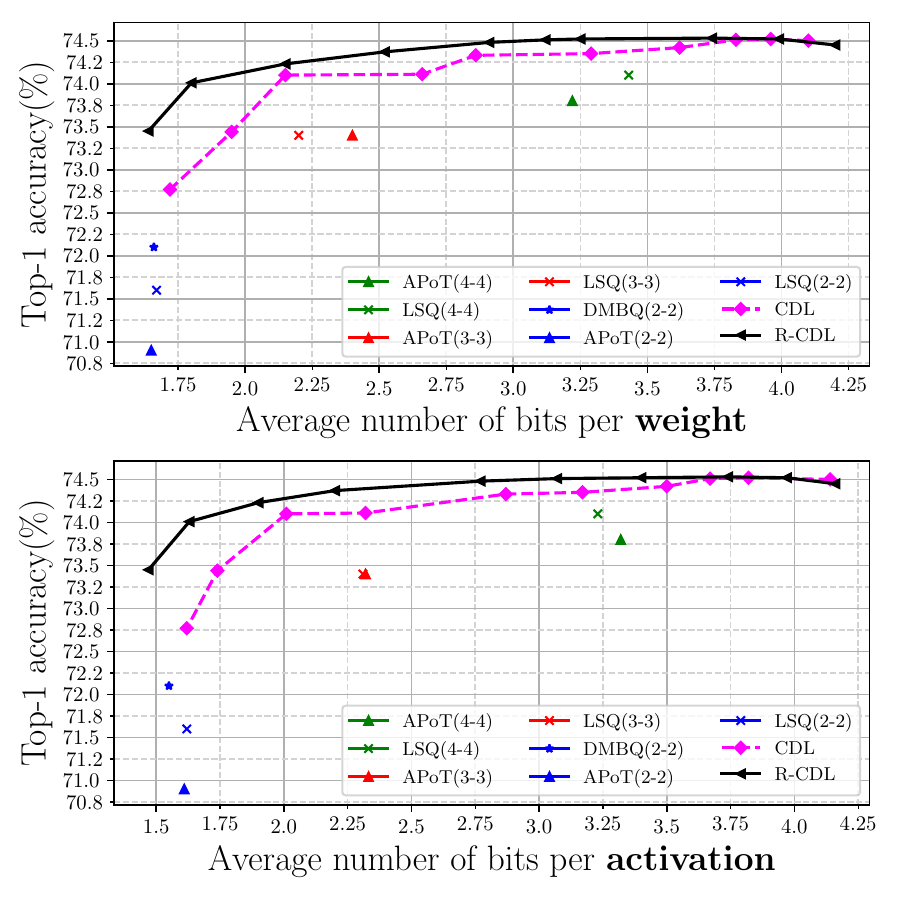}
\label{fig:ResNet-34_P}}
\vskip -0.05in
  \caption{Comparison of models trained by CDL, R-CDL, and benchmark methods in terms of the Top-1 accuracy vs the average number of bits per weight (top)/activation (bottom) on ImageNet: (a) ResNet-18, and (b) ResNet-34. All models are trained from \underline{\textbf{pre-trained}} FP models.} \label{fig:ImageNet_app}
  \vskip -0.2in 
\end{figure*}

\section{CDL with Non-uniform quantization} \label{app:non-uniform}
In this section, we demonstrate a simple way to use non-uniform quantization to further enhance the performance of R-CDL.

To this end, we use the learnable \textit{companding} quantization (LCQ) method as introduced in \cite{yamamoto2021learnable} to learn a non-uniform quantizer during training. Specifically, this non-uniform quantization is composed of three functions: 
(i) a compressing function $f_{\rm NL}(\cdot)$, (ii) a uniform quantizer $\mathsf{Q}_{\rm u}(\cdot)$, (iii) and an expanding function $f^{-1}_{\rm NL}(\cdot)$. The quantizer obtained in this manner is referred to as a \textit{companding} quantizer, and it is derived as follows:
\begin{align}
\text{comp}(\theta) = (f^{-1}_{\rm NL} \circ \mathsf{Q}_{\rm u} \circ f_{\rm NL})  (\theta),   
\end{align}
where $f_{\rm NL}(\cdot)$ is a learnable piece-wise linear function (see equation (5) in \cite{yamamoto2021learnable})  with its parameters being optimized during training. 

To integrate LCQ method into R-CDL, we replace $\mathsf{Q}_{\rm d}(\cdot)$ quantizers in R-CDL with $(f^{-1}_{\rm NL} \circ \mathsf{Q}_{\rm d} \circ f_{\rm NL})  (\theta)$ quantizers, resulting in a companding version of R-CDL referred to as R-CDL+LCQ. To show the effectiveness of R-CDL+LCQ, we have conducted experiments on ImageNet using ResNet-$\{18,34\}$ with the similar setup as those used in Section \ref{fig:non-uniform}. Additionally, we use the same hyper-parameters for $f_{\rm NL}(\cdot)$ as those employed in \cite{yamamoto2021learnable} for the ImageNet experiments. The average numbers of bits for weights and activations are still computed in the same manner as before in R-CDL, that is, after being processed by $f_{\rm NL}(\cdot)$, they are randomly quantized using the trained CPMFs and then Huffman encoded.

The results for R-CDL+LCQ and LCQ are presented in Figure \ref{fig:non-uniform}, where all models were trained from \textbf{scratch}. Based on this figure, the following observations can be made:

\noindent $\bullet$ By comparing Figs. \ref{fig:non-uniform} and \ref{fig:ImageNet}, it is evident that the performance of  R-CDL is enhanced with the implementation of non-uniform quantization.

\noindent $\bullet$ The performance of R-CDL+LCQ is superior to LCQ.

\section{Training on top of pre-trained models} \label{app:imagenet}

In this appendix, we compare CDL and R-CDL with APoT \cite{apot}, LSQ \cite{lsq} and DMBQ \cite{zhao2021distribution} when they all are applied on top of pre-trained models for ImageNet. All models in this section are trained starting from pre-trained FP models. The training settings otherwise for CDL, R-CDL, and benchmark methods are the same as those used in Subsection \ref{sec:imagenet}. 

Fig. \ref{fig:ImageNet_app} presents the performance comparison of Top-1 accuracy vs the average number of bits per weight (top)/activation (bottom) for models trained using CDL, R-CDL, and benchmark methods. Specifically, these curves are depicted for ResNet-18 and ResNet-34 in Figs. \ref{fig:ResNet-18_P} and \ref{fig:ResNet-34_P}, respectively.

Comparing Fig. \ref{fig:ImageNet_app} with Fig. \ref{fig:ImageNet} in Subsection \ref{sec:imagenet}, it is fair to make the following observations:

\noindent $\bullet$ CDL and R-CDL perform better when applied on top of pre-trained FP models compared to starting from scratch.

\noindent $\bullet$ CDL and R-CDL continue to outperform the benchmark methods when they all are applied on top of pre-trained FP models. Again in low bits, the gains offered by CDL and R-CDL, particularly by R-CDL are quite significant. 

\noindent $\bullet$ When applied on top of pre-trained FP models, CDL and R-CDL can improve the accuracy performance of the pre-trained FP models by a non-negligible margin.

\bibliographystyle{IEEEtran}
\bibliography{egbib}

\end{document}